%% file: main.tex
\documentclass[lettersize,journal]{IEEEtran}

\newif\ifrebuttal
\rebuttalfalse 

\usepackage{amsmath,amsfonts}
\usepackage{algorithm}
\usepackage{algorithmicx}
\usepackage{algpseudocode}
\DeclareMathOperator*{\argmax}{arg\,max} 
 
\usepackage{multirow}
\usepackage{amssymb}
\usepackage{pifont}
\usepackage{array}
\usepackage[caption=false,font=normalsize,labelfont=sf,textfont=sf]{subfig}
\usepackage{textcomp}
\usepackage{stfloats}
\usepackage{url}
\usepackage{verbatim}
\usepackage{graphicx}
\usepackage{enumitem}
\usepackage{xcolor}

\ifrebuttal
\else
  \colorlet{blue}{black} 
\fi

\usepackage[numbers,square,sort]{natbib}
\usepackage{makecell}
\definecolor{ForestGreen}{RGB}{34,139,34}

\definecolor{cvprblue}{rgb}{0.21,0.49,0.74}
\usepackage[
        breaklinks,
        colorlinks,
        citecolor=blue,
        urlcolor=black,
        linkcolor=blue,
        ]{hyperref}
\usepackage{multirow}

\begin{document}

\title{GCP: Guarded Collaborative Perception with Spatial-Temporal Aware Malicious Agent Detection}

\author{Yihang Tao$^*$, Senkang Hu$^{*}$, Yue Hu, Haonan An, Hangcheng Cao, and Yuguang Fang,~\IEEEmembership{Fellow,~IEEE} 

\IEEEcompsocitemizethanks{\IEEEcompsocthanksitem Y. Tao, S. Hu, H. An, H. Cao, and Y. Fang are with Department of Computer Science, City University of Hong Kong, Kowloon, Hong Kong. Email: \{yihang.tommy, senkang.forest, haonanan2-c\}@my.cityu.edu.hk, \{hangccao, my.fang\}@cityu.edu.hk. (* indicates equal contribution.)
\IEEEcompsocthanksitem   Y. Hu is with the Department of Robotics, University of Michigan, Ann Arbor, USA. Email: huyu@umich.edu.
}\thanks{
This work was supported in part by the Hong Kong Innovation and Technology Commission under InnoHK Project CIMDA,
by the Hong Kong SAR Government under the Global STEM Professorship, and by the Hong Kong Jockey Club under JC STEM Lab of Smart City (Ref.: 2023-0108).}
}

\IEEEtitleabstractindextext{%
\begin{abstract}
Collaborative perception significantly enhances autonomous driving safety by extending each vehicle's perception range through message sharing among connected and autonomous vehicles. Unfortunately, it is also vulnerable to adversarial message attacks from malicious agents, \textcolor{black}{resulting in} severe performance degradation. While existing defenses employ hypothesis-and-verification frameworks to detect malicious agents based on single-shot outliers, they overlook temporal message correlations, which can be circumvented by subtle yet harmful perturbations in model input and output spaces. This paper reveals a novel blind area confusion (BAC) attack that compromises existing single-shot outlier-based detection methods. As a countermeasure, \textcolor{black}{we propose \texttt{GCP}, a \textbf{\underline{G}}uarded \textbf{\underline{C}}ollaborative \textbf{\underline{P}}erception framework based on spatial-temporal aware malicious agent detection}, which maintains single-shot spatial consistency through a confidence-scaled spatial concordance loss, while simultaneously examining temporal anomalies by reconstructing historical bird's eye view motion flows in low-confidence regions. We also employ a joint spatial-temporal Benjamini-Hochberg test to synthesize dual-domain anomaly results for reliable malicious agent detection. Extensive experiments demonstrate \texttt{GCP}'s superior performance \textcolor{black}{under} diverse attack scenarios, achieving up to 34.69\% improvements in AP@0.5 compared to \textcolor{black}{the} state-of-the-art CP defense strategies under BAC attacks, while maintaining consistent 5-8\% improvements under other typical attacks.
        \textcolor{blue}{Code will be released at \url{https://github.com/yihangtao/GCP.git}.}
\end{abstract}

\begin{IEEEkeywords}
  Connected and autonomous vehicle (CAV), collaborative perception, malicious \textcolor{black}{agents}, spatial-temporal detection.
\end{IEEEkeywords}}

\maketitle

\IEEEdisplaynontitleabstractindextext

%
\IEEEpeerreviewmaketitle

\input{sections/introduction.tex}

\input{sections/related_work.tex}

\input{sections/problem_formulation.tex}

\input{sections/method.tex}
\input{sections/experiment.tex}

\input{sections/conclusion.tex}


%

\bibliographystyle{IEEEtran}
{
        \footnotesize
        \bibliography{ref}
        }

\input{sections/bio.tex}

\newpage
\pagestyle{empty}
\appendices
\input{sections/appendix.tex}

\end{document}

%% file: sections/introduction.tex
\section{Introduction}
\begin{figure}[t]
    \centering
    \includegraphics[width=0.95\linewidth]{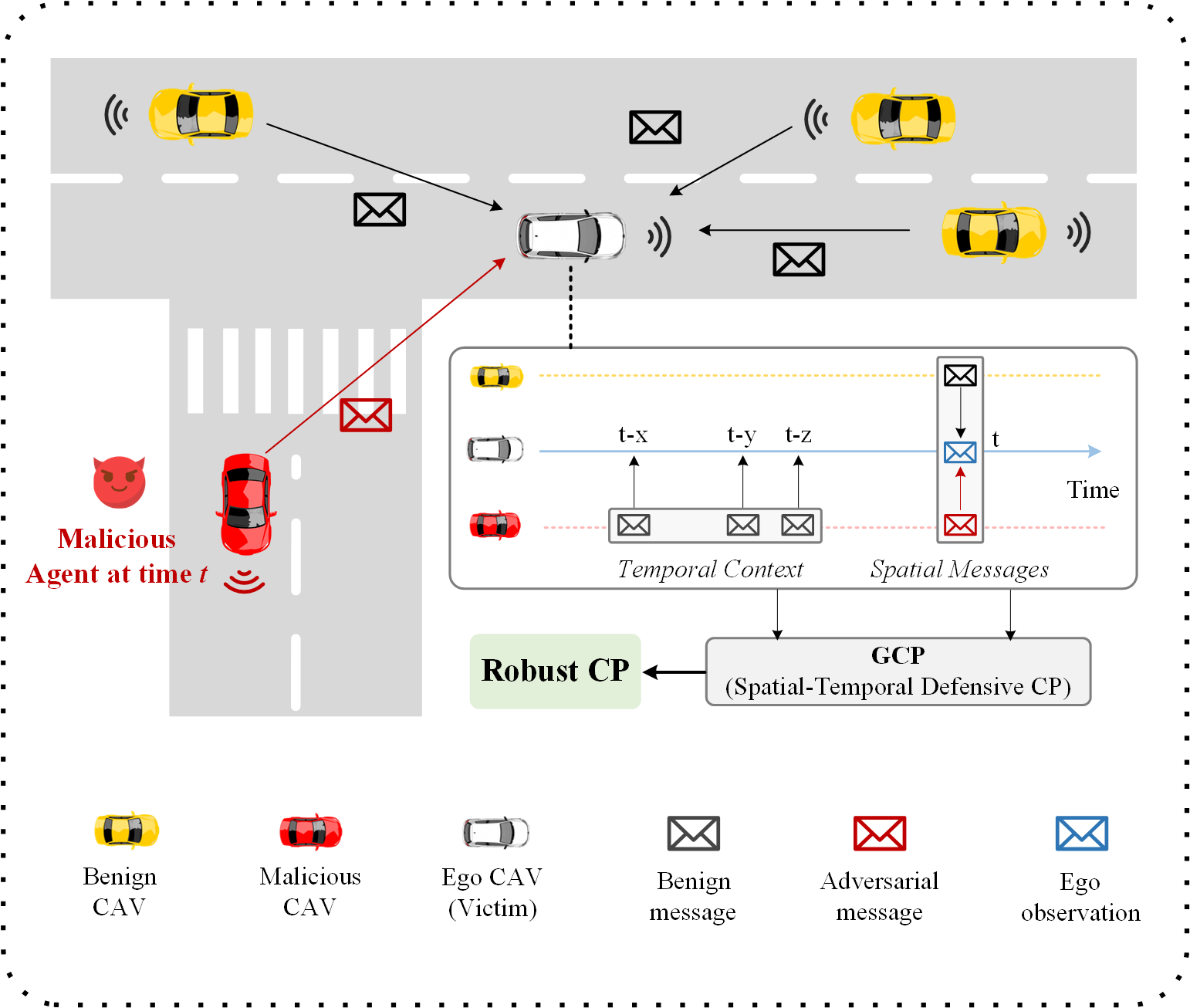}
    \caption{\textbf{Illustration of security challenges and defense mechanisms in CP.} While CP systems are vulnerable to adversarial messages from malicious agents, our proposed \texttt{GCP} framework provides comprehensive protection through joint spatial-temporal consistency verification, effectively safeguarding the system against various attack patterns.}
    \label{fig:security_threats}
    \vspace{-3mm} 
\end{figure}

\IEEEPARstart{C}{ollaborative} perception (CP) supersedes single-agent perception by enabling information-sharing among multiple connected and autonomous vehicles (CAVs), substantially enlarging a vehicle's perception scope and accuracy \cite{8885377,li2022v2x, hu2024adaptivecommunicationscollaborativeperception, hu2022wherecomm, hu2024fullscenedomaingeneralizationmultiagent,10646529,tao2024directcpdirectedcollaborativeperception, Xu_2023_CVPR}. The extended perception helps an ego vehicle detect occluded objects that were originally difficult to recognize with single-vehicle perception due to physical occlusions, thereby boosting the safety of autonomous driving. To collaborate, one simple way for an ego CAV to receive helps is to directly request early-stage raw data or late-stage detection results from the neighboring CAVs, and then combine this information with its own data to get the CP results. However, these methods are either bandwidth-consuming or vulnerable to perception noise or malicious message attacks. The recent development of deep learning has facilitated feature-level fusion, where the collaborative CAVs send an intermediate representation of deep neural network models to an ego CAV for aggregation, enhancing the performance-bandwidth trade-off in multi-agent perception.

Although CP has brought many benefits \textcolor{black}{to CP systems, it is} inevitably \textcolor{black}{attracting} adversarial attacks due to the openness of communication channels. \textcolor{black}{While traditional authentication methods (e.g., message and/or source authentication) can verify message sources and data integrity, they cannot protect against compromised legitimate agents who possess valid credentials to share malicious messages.} In an intermediate fusion-based CP system, a malicious collaborator could send an adversarial feature map with intricately crafted perturbation to an ego agent, \textcolor{black}{causing} significant CP performance degradation after fusion. This is particularly dangerous because the perturbed CP performance could drop far below the single-agent perception, consequently \textcolor{black}{resulting in} catastrophic driving decisions. Previous works have revealed diverse attacks that can fool \textcolor{black}{a} CP system \cite{294490, 9711249, Li_2023_ICCV, zhao2024maliciousagentdetectionrobust}. For example, Zhang \textit{et al.}\cite{zhao2024maliciousagentdetectionrobust} \textcolor{black}{introduced} an online attack by optimizing a perturbation on the attacker's feature map in each LiDAR cycle and reusing the perturbation over frames, which significantly lowers the performance of CP systems.

\textcolor{blue}{Existing CP defense methods are primarily unsupervised outlier detection schemes based on the hypothesize-and-verify paradigm. For instance, ROBOSAC \cite{Li_2023_ICCV} and CP-Guard \cite{hucpguard2025} employ sample consensus strategies to identify malicious agents, while MADE \cite{zhao2024maliciousagentdetectionrobust} utilizes multi-test consistency checks. Recently, CP-Guard+ \citep{anonymous2024cpguard} utilizes prior knowledge of attack patterns for training, which accelerates the detection speed yet imposes stronger assumption on the defender's knowledge. Furthermore, all these methods rely heavily on spatial information within a single slot, ignoring temporal correlations across frames. Consequently, they remain vulnerable to dynamic attacks that are spatially subtle but temporally anomalous.}

We strongly believe incorporating temporal knowledge is crucial for CP defense based on two key insights. \ding{182} First, adversarial attacks in real-world scenarios often exhibit distinct temporal patterns. When malicious agents inject perturbations intermittently to maintain stealth, these attacks manifest as anomalous variations in the temporal domain of CP results. For instance, both ROBOSAC \cite{Li_2023_ICCV} and MADE \cite{zhao2024maliciousagentdetectionrobust} observe that attackers typically alternate between sending malicious and benign messages across time slots to avoid detection. This temporal characteristic provides an additional verification dimension: beyond examining spatial consistency in the current frame, we can leverage historical clean messages from collaborators as reliable references to identify suspicious temporal deviations. \ding{183} Second, even when attackers continuously inject perturbations, their impact on CP results inevitably creates distinctive temporal patterns that differ from normal CP behaviors. These patterns manifest in various aspects, such as unnatural object motion trajectories, inconsistent detection confidence variations, or abrupt changes in spatial feature distributions across frames. Such temporal anomalies, while potentially subtle in individual frames, become more apparent when analyzed over extended long sequences.

To address the aforementioned challenges, in this paper,  we first reveal a novel adversarial attack targeted at CP, namely, the blind area confusion (BAC) attack, which generates subtle and targeted perturbation in an ego CAV's less confident areas to bypass the single-shot outlier-based detection methods. Besides, to overcome the limitations of previous malicious agent detection methods, we propose \texttt{GCP}, a defensive CP system against adversarial attackers based on knowledge from both the spatial and temporal domains. Specifically, \texttt{GCP} leverages a confidence-scaled spatial concordance loss and Long-short-term-memory autoencoder (LSTM-AE)-based BEV flow reconstruction to check the spatial and temporal consistency jointly. To sum up, our \textcolor{black}{main} contributions are three-fold:
\begin{itemize}
    \item We reveal a novel attack, dubbed as blind area confusion (BAC) attack, which is targeted at CP systems by generating subtle and dangerous perturbation in \textcolor{black}{an} an ego CAV's less confident areas. The attack can significantly degrade existing state-of-the-art single-shot outlier-based malicious agent detection methods for CP systems.
    \item We \textcolor{black}{develop} \texttt{GCP}, a novel spatial-temporal aware CP defense framework, \textcolor{black}{under which} malicious agents can be jointly detected by utilizing a confidence-scaled spatial concordance loss and an LSTM-AE-based temporal BEV flow reconstruction. To the best of our knowledge, this is the first work to protect CP systems from joint spatial and temporal views.
    \item We conduct comprehensive experiments on diverse attack scenarios with \textcolor{blue}{V2X-Sim \citep{li2022v2x}} dataset. The results demonstrate that \texttt{GCP} achieves the state-of-the-art performance, with up to 34.69\% improvements in AP@0.5 compared to \textcolor{black}{the} existing state-of-the-art defensive CP methods under intense BAC attacks, while maintaining 5-8\% advantages under other adversarial attacks targeting at CP systems.
\end{itemize}

%% file: sections/related_work.tex
\section{Related Work}
\label{sec:related_work}

\subsection{Collaborative Perception (CP)}

To overcome the inherent limitations of single-agent perception, particularly restricted field-of-view (FoV) and occlusions, collaborative perception (CP) has emerged as a promising paradigm leveraging multi-agent fusion to enhance perception accuracy \citep{hanCollaborativePerceptionAutonomous2023,huCollaborativePerceptionConnected2024}. \textcolor{blue}{The evolution of CP systems has witnessed various strategies, from early raw-data \citep{8885377} and output-level fusion \citep{10.1007/978-3-030-58589-1_10}, which faced challenges in communication overhead and information loss, to sophisticated intermediate-level feature fusion.} \textcolor{blue}{Notable advances include DiscoNet \citep{NEURIPS2021_f702defb}, which employs a teacher-student framework to learn an optimized collaboration graph, V2VNet \citep{10.1007/978-3-030-58536-5_36}, which leverages graph neural networks for efficient aggregation, and When2com \citep{Liu_2020_CVPR}, which learns to construct communication groups to decide when to communicate.} \textcolor{blue}{Where2comm \citep{hu2022wherecomm} further advances the field by introducing a confidence-aware attention mechanism that simultaneously optimizes communication efficiency and perception performance. Similarly, R-ACP \citep{Fang2025RACP} proposes a robust task-oriented communication strategy that optimizes online self-calibration and efficient feature sharing by minimizing the Age of Perceived Targets (AoPT) to ensure timely and accurate perception.} 
\textcolor{blue}{While improving performance-bandwidth trade-offs, these developments exposed critical vulnerabilities in CP systems regarding robustness against adversarial attacks, which is the main focus of this paper.} 

\subsection{Adversarial CP}

The vulnerabilities in CP systems can be broadly categorized into systematic and adversarial challenges, each presenting unique threats to system reliability and safety. On the system front, inherent issues such as communication delays, synchronization problems, and localization errors have been addressed by recent works. CoBEVFlow \citep{NEURIPS2023_5a829e29} introduced an asynchrony-robust CP system to compensate for relative motions to align perceptual information across different temporal states, while CoAlign \citep{lu2023robust} developed a comprehensive framework specifically targeting at unknown pose errors through adaptive feature alignment. 
However, beyond these system challenges lies a more insidious threat\textcolor{black}{, namely,} adversarial vulnerabilities introduced by malicious agents within the collaborative system. These adversaries can compromise system integrity by injecting subtle adversarial noise into shared intermediate representations, potentially causing catastrophic failures in critical scenarios. Initial investigations by Tu \textit{et al.} \citep{9711249} demonstrated how untargeted adversarial attacks could compromise detection accuracy in intermediate-fusion CP systems through feature perturbation. Zhang \textit{et al.} \citep{294490} advanced this research by incorporating sophisticated perturbation initialization and feature map masking techniques for more realistic, targeted attacks in real-world settings. Nevertheless, these attack methods lack sophistication in attack region selection and output perturbation constraints, making them potentially detectable by conventional defense mechanisms while highlighting the need for more robust security measures.

\subsection{Defensive CP}

In response to emerging threats, the research community has developed various defensive strategies, primarily focusing on output-level malicious agent detection through hypothesis-and-verification frameworks. \textcolor{blue}{ROBOSAC \citep{Li_2023_ICCV} and CP-Guard \citep{hucpguard2025} employ sample consensus strategies to systematically identify and exclude potentially malicious agents.} MADE \citep{zhao2024maliciousagentdetectionrobust} utilizes a multi-test framework leveraging both match loss and collaborative reconstruction loss to ensure robust consistency. Zhang \textit{et al.} \citep{294490} further utilize occupancy maps for discrepancy detection. While these unsupervised outlier-based detection methods show promise, they remain vulnerable to sophisticated attacks with subtle perturbations. \textcolor{blue}{Recently, CP-Guard+ \citep{anonymous2024cpguard} trains a feature classifier by pre-simulating attack patterns (CP-GuardBench). However, this approach relies on the strong assumption that the defender possesses prior knowledge of the attacker's perturbation mechanism to construct the training dataset. This contrasts with other methods that operate without such prior knowledge or access to simulated attack data. Furthermore, existing approaches generally overlook the temporal information, limiting their effectiveness against dynamic threats.} To address these limitations, we propose a comprehensive defense framework that considers both spatial and temporal aspects, offering a more robust solution to CP security challenges.

%% file: sections/problem_formulation.tex
\section{Attack Methodology}

\subsection{\textcolor{black}{Model} of CP}
\label{CP formulation}

Consider a scenario with $N$ CAVs, where the CAV set is denoted as $\mathcal{N}$. CAVs exchange collaboration messages with each other, and each CAV can maintain up to $K$ historical messages from its collaborators. For the $i$-th CAV, we denote its raw observation at time $t_i$ as $\mathbf{O}_i^{t_i}$, and use $\Phi_\mathtt{{enc}}^{i}$ to represent its pre-trained feature encoder that generates intermediate feature map $ \mathbf{F}^{t_i}_i = \Phi_\mathtt{{enc}}^{i}(\mathbf{O}_i^{t_i})$. The collaboration message transmitted from the $m$-th CAV to the $i$-th CAV at time $t_i$ is denoted as $\mathbf{F}_{m\rightarrow i}^{t_i}$. Based on the received messages at current timestamp and historical $k\ (0\le k \le K)$ timestamps, the $i$-th CAV uses a feature aggregator $f_\mathtt{{agg}}^{i}(\cdot)$ to fuse these feature maps and adopts a feature decoder  $\Phi_\mathtt{{dec}}^{i}(\cdot)$ to get the final CP output, which is expressed as:
\begin{equation}
     \mathbf{Y}_i^{t_i} = \Phi_\mathtt{{dec}}^{i}\left(f_\mathtt{{agg}}^{i}\left(\{\mathbf{F}_{m\rightarrow i}^{t_i}, \mathbf{F}_{m\rightarrow i}^{t_{i-1}}, \cdots, \mathbf{F}_{m\rightarrow i}^{t_{i-k}}\}_{m=1}^{N}\right)\right),
     \label{formulation}
\end{equation}
where $\mathbf{Y}_i^{t_i}$ is the CP result of the $i$-th CAV at time $t_i$. There are two important notes in terms of the formulation described in Eq. \ref{formulation}: i) the times $t_{i-1}, t_{i-2}, \cdots, t_{i-k}$ are not necessary to be equally distributed, and the time intervals between two consecutive times can be irregular. ii) When the length of historical times $k$ equals 0, the $i$-th CAV outputs the CP results without referring to the temporal contexts of messages of collaborative CAVs, which degrades to the settings used in most existing works like ROBOSAC \citep{Li_2023_ICCV} and MADE \citep{zhao2024maliciousagentdetectionrobust}.

\textcolor{blue}{However, this collaborative process is vulnerable when malicious agents $\mathcal{M}_{t_i} \subseteq \mathcal{N}$ exist. These agents can encode raw observations into initial feature maps and then generate adversarial perturbations $\delta$ by maximizing the distance between CP results and ground truth (GT):}
\begin{equation}
    \argmax_{\delta} \mathcal{L}(\mathbf{Y}_\delta^{t_j}, \mathbf{Y}_\mathtt{gt}^{t_j}), \quad \mathtt{s.t.}\quad  \|\delta\|\leq \Delta_i,
\end{equation}
\textcolor{blue}{where $\mathcal{L}$ is the perception loss function, $\mathbf{Y}_\delta^{t_j}$ and $\mathbf{Y}_\mathtt{gt}^{t_j}$ denote the perturbed CP output and the ground truth at time $t_j$, respectively, and $\Delta_i$ is the maximum allowable perturbation amplitude. Finally, they transmit these perturbed features to the victim to degrade perception accuracy.}

\subsection{\textcolor{blue}{Adversarial Threat Model}}

\textcolor{blue}{We adopt the same threat model assumptions as in previous works \citep{Li_2023_ICCV, zhao2024maliciousagentdetectionrobust, hucpguard2025}. Specifically, we consider an internal attacker who has compromised at least one legitimate CAV and gained white-box access to the shared CP model architecture, parameters, and the feature maps transmitted among vehicles. This access is inherent to the compromised vehicle's legitimate participation in the CP system. The attacker aims to mislead the victim's perception by sending malicious feature maps while maintaining stealthiness. We also assume the attacker adheres to the system's communication protocol regarding timing and does not manipulate artificial delays or jitter, focusing instead on manipulating the semantic content of the transmitted features.}

\begin{figure}[t]
    \centering
    \includegraphics[width=0.95\linewidth]{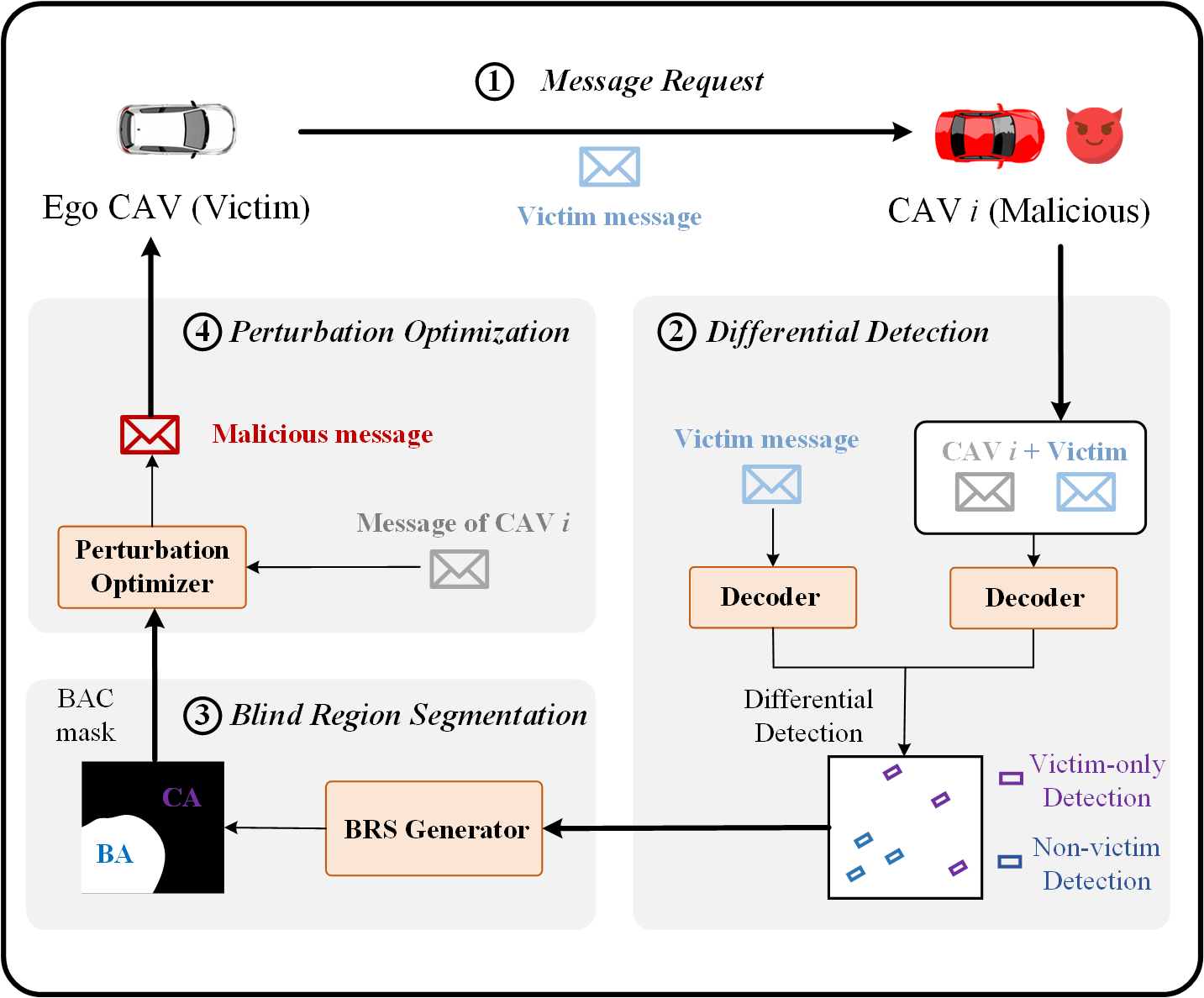}
    \caption{\textbf{Overview of the proposed blind area confusion (BAC) attack.} The malicious agent first establishes communication with the victim ego CAV to obtain collaborative messages, then infers the victim's blind regions through differential detection analysis and region segmentation. Finally, it generates adversarial perturbations guided by the inferred confidence mask to confuse the victim's perception defense system.}
    \label{fig:attack}
    \vspace{-3mm} 
\end{figure}

\begin{algorithm}[t]
    \small
    \caption{Blind Region Segmentation \textcolor{black}{(BRS)}}
    \label{alg:seg}
    \textbf{Input}: $\mathbf{M}_{c}^{t_{k}}$ (initial confidence mask), $\mathbf{Y}_{vic}^{t_{k}}$ (victim detections), $\mathbf{Y}_{nvic}^{t_{k}}$ (unseen detections), $\mathbf{D}_{i}^{t_{k}}$ (BEV detection map)
    
    \textbf{Output}: $\mathbf{M}_{c}^{t_{k}}$ (binary confidence mask)
    \begin{algorithmic}[1] 
    \Procedure{\texttt{BRS}}{$\mathbf{D}_{i}^{t_{k}}$}
        \State $e \leftarrow Cluster(\mathbf{Y}_{vic}^{t_{k}})$ \Comment{\texttt{Victim Grid}}
        \If{$\mathbf{Y}_{nvic}^{t_{k}} = \emptyset$}
            \State $\mathbf{Y}_{nvic}^{t_{k}} \leftarrow \mathop{\arg\max}\limits_{g \in \mathbf{D}_{i}^{t_{k}}} Dist(g,e)$
        \EndIf
        \State $\mathbf{M}_{c}^{t_{k}}\{\mathbf{D}_{i}^{t_{k}}\{\mathbf{Y}_{vic}^{t_{k}}\}\} \leftarrow 1$; $\mathbf{M}_{c}^{t_{k}}\{\mathbf{D}_{i}^{t_{k}}\{\mathbf{Y}_{nvic}^{t_{k}}\}\} \leftarrow -1$ 
        \State $\mathbf{Q}_{ca}, \mathbf{Q}_{ba} \leftarrow Queue\{m\}, \forall\ \mathbf{M}_{c}^{t_{k}}\{m\} = 1\ \text{or}\ -1$
        \While{$\mathbf{Q}_{ca} \neq \emptyset$ \textbf{or} $\mathbf{Q}_{ba} \neq \emptyset$} \Comment{\texttt{Region Growing}}
            \For{$\mathbf{Q} \in \{\mathbf{Q}_{ca}, \mathbf{Q}_{ba}\}$ \textbf{and} $\mathbf{Q} \neq \emptyset$}
                \State $s \leftarrow \mathbf{Q}.Pop()$
                \For{$j \in GetNeighbors(s, e)$ \textbf{and} $\mathbf{M}_{c}^{t_{k}}\{j\} = 0$}
                    \State $\mathbf{M}_{c}^{t_{k}}\{j\} \leftarrow [\mathbf{Q} = \mathbf{Q}_{ca}] ? 1 : -1$
                    \State $\mathbf{Q}.Append(j)$
                \EndFor
            \EndFor
        \EndWhile
        \State $\mathbf{M}_{c}^{t_{k}}\{m\} \leftarrow 0, \forall\ \mathbf{M}_{c}^{t_{k}}\{m\} = -1$ \Comment{\texttt{Binary Mask}}
        \State \Return $\mathbf{M}_{c}^{t_{k}}$
    \EndProcedure
    \end{algorithmic}
\end{algorithm}

\subsection{Blind Area Confusion Attack}
\label{bsi}

For previous outlier-based methods like ROBOSAC \citep{Li_2023_ICCV} and MADE \citep{zhao2024maliciousagentdetectionrobust}, the core idea is to check the spatial consistency between ego CAV and CP outcomes. However, they only clip the perturbation at the input level, and the output detection errors are significant and randomly distributed, which can be easily detected by outlier-based defense methods. Intuitively, we want to emphasize that a good attack targeted at the CP system should satisfy two conditions: (i) both the input and output perturbation should not exceed a certain level; (ii) the output perturbation should be more distributed in the regions where the victim vehicle (i.e., agent or CAV) is less sensitive so that the victim vehicle can hardly discern whether the unseen detections \textcolor{black}{could} benefit from collaboration or just fake ones.
Following the above ideas, we design a novel blind area confusion (BAC) attack as shown in Figure \ref{fig:attack}. The steps are elaborated below.

    \textbf{Message Request.} Following our threat model, a malicious CAV $i$ can participate in the CP system before being identified. During this initial phase, it masquerades as a benign agent to establish communication with the victim's ego vehicle. Specifically, the malicious CAV requests feature messages $\mathbf{F}_{e \rightarrow i}^{t_{k}}$ from the victim's ego agent at timestamp $t_k$. These messages contain valuable information about the victim's perception capabilities and limitations, which will be exploited in subsequent attack stages. Note that in our low frame rate setting (e.g., 10 FPS), the attack generation and feature fusion can be completed within the same frame, eliminating the need for temporal prediction compensation that would be required in high FPS scenarios.
    
    \textbf{Differential Detection.} After obtaining \textcolor{black}{a} victim's messages, the malicious CAV $i$ performs a comparative analysis between independent and CP results to infer the victim's blind spots. The malicious CAV first generates two types of perception results: single perception $\mathbf{Y}_{s}^{t_{k}} = \Phi_\mathtt{{dec}}^{i}\left(\mathbf{F}_{e \rightarrow i}^{t_{k}}\right)$ using only the victim's transmitted features, and CP $\mathbf{Y}_{c}^{t_{k}} = \Phi_\mathtt{{dec}}^{i}\left(f_\mathtt{{agg}}^{i}\left(\mathbf{F}_{i\rightarrow i}^{t_{k}}, \mathbf{F}_{e\rightarrow i}^{t_{k}}\right)\right)$ using both local and \textcolor{black}{the} victim's features. Let $\mathbf{Y}_{s}^{t_{k}} \cap \mathbf{Y}_{c}^{t_{k}}$ denote the matched bounding boxes between single and CP results. Based on this matching, the system partitions all detections into two categories: victim-only detections $\mathbf{Y}_{vic}^{t_{k}} = \mathbf{Y}_{s}^{t_{k}}$ (marked in \textcolor{violet}{purple}), and non-victim detections $\mathbf{Y}_{nvic}^{t_{k}} = \mathbf{Y}_{c}^{t_{k}} \setminus (\mathbf{Y}_{s}^{t_{k}} \cap \mathbf{Y}_{c}^{t_{k}})$ (marked in \textcolor{blue}{blue}). This differential detection process reveals the spatial distribution of the victim's unique detections, providing crucial insights into its perception strengths and potential blind spots, which form the foundation for subsequent targeted perturbation generation.
    
    \textcolor{black}{
    \textbf{Blind Region Segmentation.} The differential detection map is processed through an adaptive region growing algorithm to partition the BEV detection map into confident area (CA) and blind area (BA), as shown in Algorithm \ref{alg:seg}. The algorithm first uses $Cluster()$ to determine the victim grid $e$ by finding the grid point with the minimum total distance to all victim-detected objects, establishing a perception-centric coordinate system. It then initializes seed grids from $\mathbf{Y}_{vic}^{t_{k}}$ as CA seeds (value 1) and $\mathbf{Y}_{nvic}^{t_{k}}$ as BA seeds (value -1). When $\mathbf{Y}_{nvic}^{t_{k}}$ is empty, which occurs in limited perception range scenarios, the grid farthest from $e$ is selected as the BA seed to reflect natural perception degradation with distance.
    The region growing process utilizes two priority queues ($\mathbf{Q}_{ca}$, $\mathbf{Q}_{ba}$) to manage confident and blind area expansion. The $GetNeighbors()$ function implements an adaptive neighbor selection mechanism:
    \begin{equation}
        \label{ks}
            K_s = \left\lceil K_{base}\cdot\exp\left(-\gamma_d\cdot\frac{Dist(s,e)}{D_{norm}}\right)\right\rceil,
    \end{equation}
    where $Dist(s,e)$ computes the Euclidean distance between grid $s$ and victim grid $e$, $D_{norm}=\sqrt{H^2+W^2}$ is the normalization factor based on BEV detection map dimensions (height $H$ and width $W$), $K_{base}=6$ establishes a hexagonal-like growth pattern, and $\gamma_d=0.3$ controls the decay rate. This distance-adaptive design ensures denser expansion near the victim and sparser expansion in distant regions, naturally modeling spatial perception reliability \citep{9156791}.
    \textcolor{blue}{The expansion proceeds simultaneously via queue operations. Initialized by $Queue\{m\}$, priority queues $\mathbf{Q}_{ca}$ and $\mathbf{Q}_{ba}$ drive breadth-first growth. Grids are iteratively processed using $Queue.Pop()$, where unassigned neighbors inherit the label (1 for CA, -1 for BA) and are added via $Queue.Append()$. Finally, converting BA labels to 0 generates the binary mask $\mathbf{M}_{c}^{t_{k}}$, which effectively delineates the victim's perception boundaries.}
    }

    \textbf{Perturbation Optimization.} BAC perturbation optimization aims to make the perturbation mostly distributed in the victim-blind area of the BEV detection map while adaptively controlling the perturbation magnitude. This optimization problem can be formulated as:
    \begin{equation}
        \begin{aligned}
         \argmax_{\delta_{bsi}}\quad &\mathcal{L}(\mathbf{Y}_\delta^{t_{k}} \odot \mathbf{W}_{\delta}, \mathbf{Y}_\mathtt{gt}^{t_{k}} \odot \mathbf{W}_{\delta}), \\
         \mathtt{s.t.}\quad    & \mathbf{W}_{\delta} = w_{\delta} \mathbf{M}_{\delta} + \mathbf{S}_{\delta}, \\
         & \mathbf{S}_{\delta} = 1 - \sigma\left(|\mathbf{Y}_\delta^{t_{k}} - \mathbf{Y}_\mathtt{gt}^{t_{k}}| - \Delta_{o}\right), \\
         & \|\delta_{bsi}\|\leq \Delta_i,\\
        \end{aligned}
    \end{equation}
    where $\mathcal{L}(\cdot)$ denotes the optimization loss function, $\mathbf{M}_{\delta} = \mathbf{1} -\mathbf{M}_{c}^{t_{k}}\{\mathbf{D}_{i}^{t_{k}}\{\mathbf{Y}_{\delta}^{t_{k}}\}\}$ is the inverted confidence mask, $\sigma(\cdot)$ is the sigmoid activation function, $|\cdot|$ computes element-wise absolute values, $\Delta_i$ bounds the input perturbation magnitude, $\Delta_{o}$ bounds the output perturbation magnitude, \textcolor{black}{and} $w_{\delta}$ is a positive weighting parameter.
    
    This optimization formulation incorporates several innovative design principles for effective and stealthy attacks. First, instead of a simple loss function, it employs a weighted loss function where $\odot$ denotes element-wise multiplication. The weight $\mathbf{W}_{\delta}$ is carefully designed to guide the spatial distribution of perturbations, ensuring more targeted attacks. Second, the optimization adopts a dual-weight mechanism: the spatial guidance weight $\mathbf{M}_{\delta}$ directs perturbations towards the victim's blind areas, while the adaptive suppression weight $\mathbf{S}_{\delta}$ automatically reduces weights when output perturbations become too large. The adaptive suppression is implemented through a sigmoid function, which provides smooth transitions when the prediction deviates from the ground truth beyond the threshold $\Delta_{o}$. This design ensures that the attack remains effective while avoiding generating easily detectable anomalies. Additionally, the input perturbation magnitude is constrained by $\Delta_i$ to maintain physical feasibility and attack stealthiness.
    After the optimization, the malicious CAV $i$ incorporates the optimized perturbation $\delta_{bsi}$ into its intermediate BEV feature before transmission to the victim CAV.

%% file: sections/method.tex
\begin{figure*}[t]
    \centering
    \includegraphics[width=0.95\linewidth]{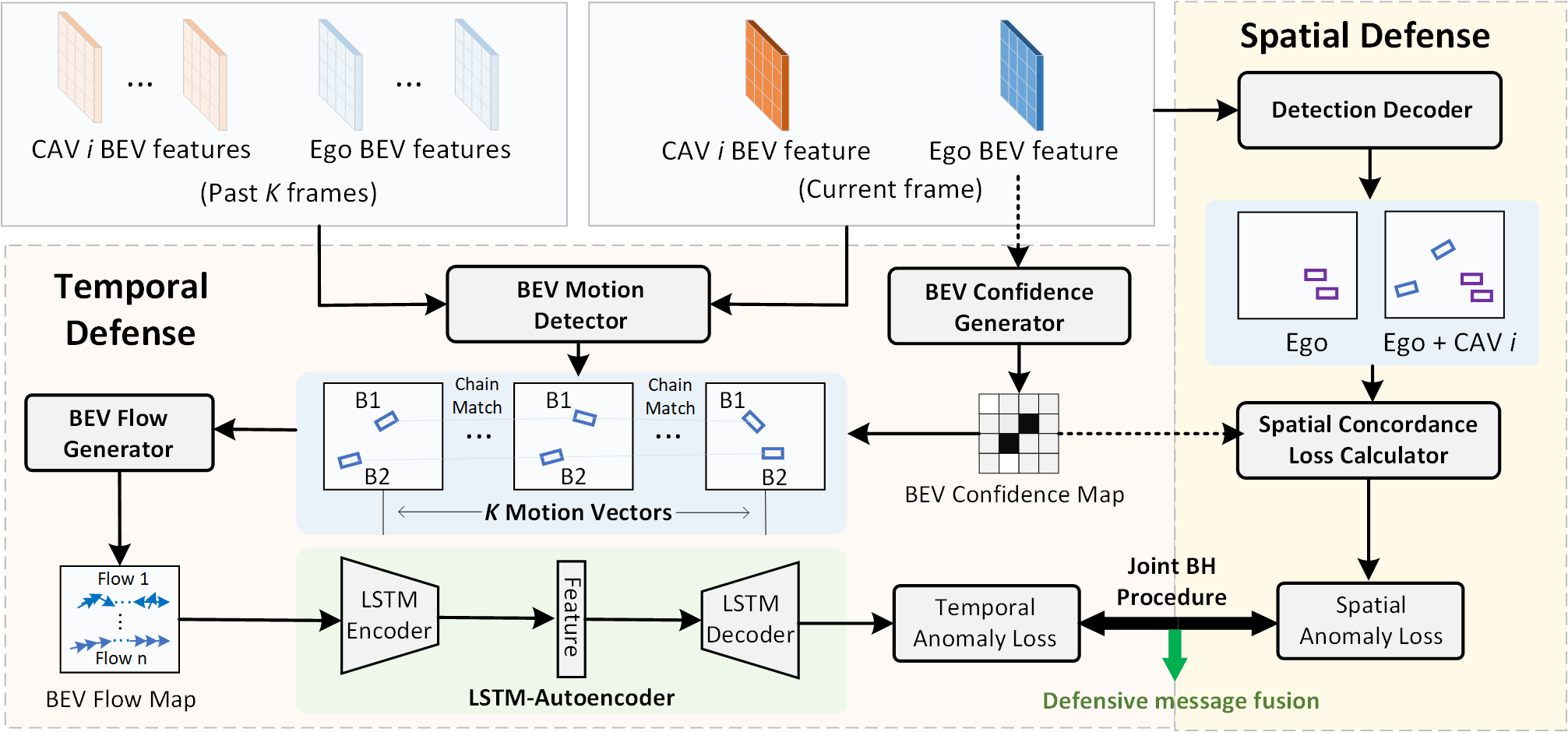}
    \caption{\textbf{Overview of the proposed \texttt{GCP} framework.} \texttt{GCP} performs joint spatial-temporal consistency verification through two key components: (1) a confidence-scaled spatial concordance loss that adaptively evaluates detection consistency, and (2) an LSTM-AE-based temporal BEV flow reconstruction that captures motion patterns in CP.}
    \label{fig:overall}
    \vspace{-3mm} 
\end{figure*}

\section{GCP Framework}

\subsection{Overall Architecture}
\label{overall}
In this paper, we propose a robust framework to guard CP systems through spatial-temporal aware malicious agent detection. 
As illustrated in Fig. \ref{fig:overall}, \texttt{GCP} operates through a dual-domain verification process. First, it computes a confidence-scaled spatial concordance loss by estimating confidence scores for each grid cell in the BEV detection map, evaluating the consistency between the ego CAV's observations and messages from the $i$-th neighboring CAV at the current time slot. Second, it performs temporal verification by analyzing the past $K$ frames from both the ego CAV and the $i$-th neighboring CAV to generate a BEV flow map for low-confidence detections, employing an LSTM-AE-based temporal reconstruction to verify motion consistency. The framework culminates in a comprehensive spatial-temporal multi-test that combines both consistency metrics to make final decisions on malicious agent detection. The following sections (\ref{cwml} to \ref{kf}) provide the detailed descriptions of these core components.

\subsection{Confidence-Scaled Spatial Concordance Loss}
\label{cwml}

Existing spatial consistency checking methods typically compare the difference between the ego CAV's perception and CP without considering the ego CAV's varying perception reliability across different spatial regions. This indiscrimination \textcolor{black}{could result in} mistaking true blindspot detection complemented by other CAVs as malicious signals. 
\textcolor{black}{This occurs because the ego CAV's perception reliability naturally decreases in occluded areas and at sensor range boundaries \citep{9156791}. When other CAVs provide valid detections in these low-reliability regions, traditional methods may incorrectly flag them as inconsistencies by failing to account for the ego CAV's spatially varying perception capabilities.}
To address this limitation, we propose a novel confidence-scaled spatial concordance loss (CSCLoss) that incorporates the ego CAV's detection confidence when checking messages from the $i$-th CAV at time $t_k$.
Let $\mathbf{Y}_{e}^{t_{k}}$ denote the ego CAV's detected bounding boxes and $\mathbf{Y}_{e,i}^{t_{k}}$ represent the collaboratively detected bounding boxes at timestamp $t_{k}$. We first construct a weighted bipartite graph between these two sets, where edges represent matching costs based on classification confidence and intersection-over-union (IoU) scores. Since $|\mathbf{Y}_{e}^{t_{k}}|$ may differ from $|\mathbf{Y}_{e,i}^{t_{k}}|$, we pad the smaller set with empty boxes to ensure the one-to-one matching. The optimal matching $\mathcal{O}\{\mathbf{Y}_{e}^{t_{k}}, \mathbf{Y}_{e,i}^{t_{k}}\}_{n=1}^{N}$ is then obtained using the Kuhn-Munkres algorithm \citep{8685190}.
To compute the CSCLoss, we first estimate a spatial confidence map $\mathbf{C}_{e}^{t_{k}}$ using the ego CAV's feature map $\mathbf{F}_{e\rightarrow e}^{t_{i}}$:
\begin{equation}
    \mathbf{C}_{e}^{t_{k}} = \Phi_{conf}\left(\mathbf{F}_{e\rightarrow e}^{t_{i}}\right) \in [0,1]^{H\times W},
\end{equation}
where $\Phi_{conf}(\cdot)$ is a detection decoder that generates confidence scores for each grid cell in the BEV map \citep{hu2022wherecomm}.
The CSCLoss is then calculated by combining the optimal matching and confidence map:
\begin{equation}
    \mathcal{L}_{csc} \left(\mathbf{Y}_{e}^{t_{k}}, \mathbf{Y}_{e,i}^{t_{k}}\right) = \sum_{c \in \mathcal{C}}\sum_{j \in \mathcal{O}}{\frac{\mathcal{M}\left(\mathbf{Y}_{e}^{t_{k}(j)}, \mathbf{Y}_{e,i}^{t_{k}(j)}; c\right)\cdot\mathbf{C}_{e}^{t_{k}(j)}}{|\mathbf{C}_{e}^{t_{k}}|}},
\end{equation}
where
$\mathbf{C}_{e}^{t_{k}(j)}$ is the confidence score for the grid cell containing bounding box $\mathbf{Y}_{e}^{t_{k}(j)}$,
$c \in \mathcal{C}$ represents a prediction class,
$\mathcal{M}(\cdot)$ computes the matching cost between two boxes:
\begin{equation}
    \mathcal{M}\left(y_1, y_2; c\right) = \operatorname{ReLU} \left(p_{1} - p_{2}\right) + \phi\left(1-\operatorname{IoU}\left(z_{1}, z_{2}\right)\right),
\end{equation}
where $p_{1}$ and $p_{2}$ are the class $c$ posterior probabilities for boxes $y_1$ and $y_2$, \textcolor{black}{respectively}, $z_{1}$ and $z_{2}$ are their spatial coordinates, and $\phi$ is a weighting coefficient. Typically, when temporal context is not applicable, a spatial anomaly can be assumed to be detected when $\mathcal{L}_{csc}$ exceeds a threshold $\alpha$.

\subsection{Low-Confidence BEV Flow Matching}
\label{flow}

Bird's Eye View (BEV) flow \citep{NEURIPS2023_5a829e29} represents the consecutive motion vectors of detected objects across consecutive frames in the top-down perspective. By observing the flow of bounding boxes in the BEV space over time, we can capture the temporal dynamics and motion patterns of objects in the scene.
For computational efficiency, we only perform temporal matching on two specific types of low-confidence BEV flows: (i) those with low detection scores from ego CAV's own view, and (ii) ego CAV's unseen detections that the collaborative agents complement. While temporal matching could be applied to all detected boxes, focusing on these low-confidence cases is particularly crucial as \textcolor{black}{it is} difficult for an ego CAV to judge whether they represent true objects or perturbations caused by malicious agents merely through single-shot spatial checks, especially when the added perturbation is subtle. 
To address this challenge, we utilize temporal characteristics to analyze their anomaly patterns.
We first match the detected low-confidence bounding boxes of $\mathbf{Y}_{e,i}^{t_{k}}$ to those in the past $K$ frames $\mathbf{Y}_{e,i}^{t_{k-1}}, \mathbf{Y}_{e,i}^{t_{k-K+1}}, \cdots, \mathbf{Y}_{e,i}^{t_{k-K}}$. 
In the BEV detection map, each detected bounding box can be represented as a set of 4 corner points:
\begin{equation}
    o_j = [x_j^1,y_j^1, \cdots, x_j^4, y_j^4]^\top \in \mathbb{R}^8, \forall o_j \in  \mathbf{Y}_{e,i}^{t_{k}},
\end{equation}
where $\{(x_j^k, y_j^k)\}_{k=1}^{4}$ represents the locations of 4 corners of bounding box $o_j$ in BEV detection map.
Given the current-frame low-confidence bounding boxes set $\mathbf{O}_{e,i}^{t_{k}} \in \mathbf{Y}_{e,i}^{t_{k}}$, the ego CAV will generate BEV flow $\mathbf{I}_{e,i}^{t_{k}}$ by iteratively matching the bounding boxes in each historical frame.
As shown in Algorithm \ref{alg:GCP}, the matching process is a chain-based process, which means \textcolor{black}{that} we first find the best-matched bounding boxes $\mathbf{B}_{e,i}^{t_{k-1}}$ \textcolor{black}{in} $\mathbf{Y}_{e,i}^{t_{k-1}}$ for $\mathbf{O}_{e,i}^{t_{k}}$, then we keep searching for the best matched bounding boxes $\mathbf{B}_{e,i}^{t_{k-2}}$ \textcolor{black}{in} $\mathbf{Y}_{e,i}^{t_{k-2}}$ for $\mathbf{B}_{e,i}^{t_{k-1}}$, until the $K$-th frame has been matched.
During this process, not all the bounding boxes in $\mathbf{O}_{e,i}^{t_{k}}$ \textcolor{black}{in} $\mathbf{Y}_{e,i}^{t_{k}}$ can be matched up to the $K$-th frame, we call the successfully matched bounding boxes as the candidate BEV flow $\mathbf{I}_{c}^{t_{k}}$. LSTM-AE is used to reconstruct it for further temporal consistency check. As for the unmatched BEV flow $\mathbf{I}_{u}^{t_{k}}$, they will \textcolor{black}{result in} additional time anomaly penalties.
To maintain computational efficiency, we cache the historical matching chains for each tracked flow. This significantly reduces the computational complexity as previously established matches can be directly reused without re-computation, making the chain matching process highly efficient in practice.
Note that consecutive $K$ frames are not always completely available for chain matching. There are two cases when the frame cache is not enough for chain matching. 

\ding{182} \textit{Case 1:} The ego \textcolor{black}{CAV} may have not collected up to $K$ frames of neighboring CAVs at the early stage. In this case, we only use spatial consistency to check the malicious agent until the cached messages are up to $K$ frames.

\ding{183} \textit{Case 2:} Certain frames of neighboring CAVs have been identified as malicious messages and are thereby discarded. In this case, we use Kalman Filter (KF) \citep{10517905} for interpolation (Please refer to Appendix \ref{appendix:Kalman}). We also set a maximum consecutive interpolation limit of $L$ to avoid the cumulative error.
Once the consecutive interpolation frames exceed the limit $L$, all the cached frames will be refreshed.

\subsection{Temporal BEV Flow Reconstruction}
\label{kf}

For the generated candidate BEV flow, we further analyze their temporal characteristics based on the current and past $K$-frame messages from the $i$-th CAV. As shown in Fig \ref{fig:overall}, there are three key components of BEV flow reconstruction: LSTM encoder,  LSTM decoder, and temporal reconstruction loss estimator. 

\textbf{LSTM encoder.} Each candidate BEV flow $\mathbf{i}_{c}^{t_k} \in \mathbf{I}_{c}^{t_k}$ can be represented as $\mathbf{i}_{c}^{t_k} = [o_{k-K}, o_{k-K+1}, \cdots, o_{k}]^\top \in \mathbb{R}^{(K + 1) \times 8}$, the LSTM encoder further encodes the high-dimensional input sequence into a low-dimensional hidden representation using the following equation:
\begin{equation}
\label{lstm1}
    H_k = \sigma_o\left([H_{k-1}, o_k]\right)\odot \tanh(C_k), 
\end{equation}
where $\sigma_o(\cdot) = \sigma(w_o(\cdot) + b_o)$ represents the output gate with activation function $\sigma(\cdot)$,  weight $w_o$, and bias $b_o$. $H_{k-1}$ and $o_k$ represent the concatenation of the hidden state and the current input, respectively. $C_k$ represents the input gate calculated by:
\begin{equation}
\label{lstm2}
    \hat{C}_k = \tanh(w_c[H_{k-1}, o_k] + b_c),
\end{equation}
\begin{equation}
\label{lstm3}
    C_k =  \sigma_f\left([H_{k-1}, o_k]\right)\odot C_{k-1} + \sigma_g\left([H_{k-1}, o_k]\right)\odot \hat{C}_k,
\end{equation}
where $\sigma_f(\cdot)$ is the forget gate, $\sigma_g(\cdot)$, $\hat{C}_k$, $C_k$ represent the input gate. The output vector will be repeated $K + 1$ times to yield the final encoded feature vector:
\begin{equation}
    H_k = \Phi_{enc}^{lstm} \{\mathbf{i}_{c}^{t_k}\} \in \mathbb{R}^{1 \times M},
\end{equation}
\begin{equation}
    \mathbf{H}_{bf} = \overbrace{H_k \oplus H_k \oplus \ldots \oplus H_k}^{K + 1\ \text{times}}  \in \mathbb{R}^{(K + 1) \times M},
\end{equation}
where $M$ is the latent encoded feature dimension, $\Phi_{enc}^{lstm}\{\cdot\}$ is the LSTM encoding network model containing $M$ computation units following Eq. \ref{lstm1} to Eq. \ref{lstm3}, and $\oplus$ represents the concatenation operation along the horizontal dimension.

\textbf{LSTM decoder.} The LSTM Decoder consists of a 3-layer network with $K$ LSTM cell units. Each LSTM cell processes each $\mathbb{R}^{1\times M}$ encoded feature. These LSTM units generate a $\mathbb{R}^{(K + 1)\times M}$ output vector learned from the encoded feature, which is further multiplied with a $\mathbb{R}^{M \times 8}$ vector output by a TimeDistributed (TD) layer \citep{wei2023reconstructionbasedlstmautoencoderanomalybasedddos}. The TimeDistributed layer maintains the temporal structure by applying a fully connected layer to each time step output. Finally, the LSTM decoder generates a reconstructed vector $\hat{\mathbf{i}}_{c}^{t_k}$ with the same size as the input vector following:
\begin{equation}
    \hat{\mathbf{i}}_{c}^{t_k} = \Phi_{dec}^{lstm}\{\mathbf{H}_{bf}\}\cdot\operatorname{\textbf{TD}}\{\mathbf{H}_{bf}\}  \in \mathbb{R}^{(K + 1)\times 8},
\end{equation}
where  $\Phi_{dec}^{lstm}\{\cdot\}$ is the LSTM decoding network model, $\operatorname{\textbf{TD}}\{\cdot\}$ represents the TimeDistributed Layer funciton.

\textbf{Temporal reconstruction loss estimator.} To evaluate the loss between the input vector and output vector, we use Mean Absolute Error (MAE) as a metric, given by:
\begin{equation}
    \mathcal{L}_{tr}(\mathbf{i}_{c}^{t_k}, \hat{\mathbf{i}}_{c}^{t_k}) = \frac{1}{K} \sum_{i=1}^{K} |o_{k-i} - \hat{o}_{k-i}|,
\end{equation}
where $\hat{o}_{k-i} \in \hat{\mathbf{i}}_{c}^{t_k}$ is the reconstructed BEV flow vector at time $t_{k-i}$. 
The final temporal anomaly score is the sum of the candidate BEV flow reconstruction loss and the unmatched BEV flow penalty, \textcolor{black}{given} by:
\begin{equation}
    \mathcal{L}_{ta} = \sum_{\mathbf{i}_{c}^{t_k} \in \mathbf{I}_{c}^{t_k}} \mathcal{L}_{tr}(\mathbf{i}_{c}^{t_k}, \hat{\mathbf{i}}_{c}^{t_k}) + \kappa_p |\mathbf{I}_{u}^{t_k}|,
\end{equation}
where $\kappa_p$ is a constant penalty coefficient for the unmatched BEV flow.

\begin{figure}[t]
    \centering
    \includegraphics[width=0.95\linewidth]{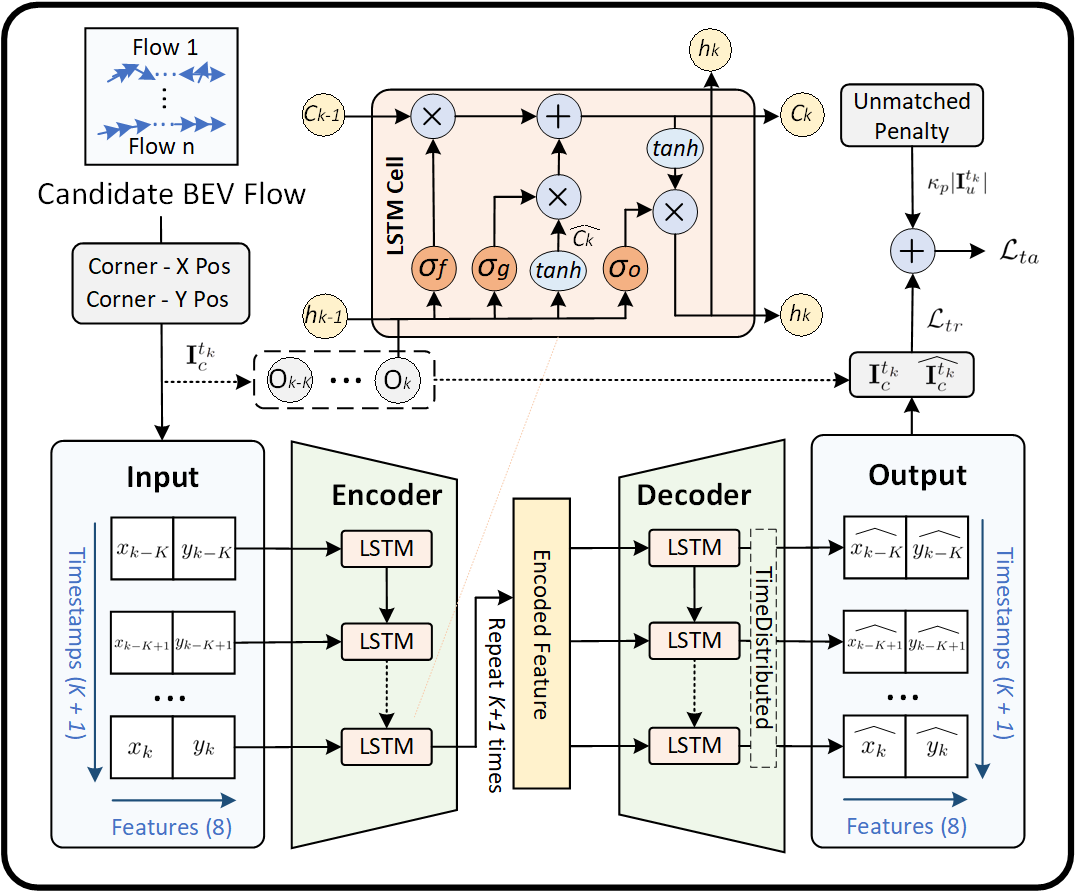}
    \caption{\textbf{Architecture of LSTM-AE-based BEV flow reconstruction.} The input BEV flow vector consists of 8-dimensional features representing corner points of detected objects. The encoded latent features are repeated $K + 1$ times before decoding, followed by a TimeDistributed layer for temporal-aware reconstruction of object motion patterns.}
    \label{fig:lstm-ae}
    \vspace{-3mm} 
\end{figure}

\subsection{Joint Spatial-Temporal Benjamini-Hochberg Test}

To jointly utilize spatial and temporal anomaly results for malicious agent detection, we employ the Benjamini-Hochberg (BH) procedure \cite{benjamini1995controlling, zhao2024maliciousagentdetectionrobust}. The BH procedure effectively controls the False Discovery Rate (FDR), the expected proportion of false positives among all rejected null hypotheses, which is crucial in CP where false accusations can severely impact system reliability.
For each CAV $i$ under inspection, we first compute a weighted combination of spatial and temporal scores:
\begin{equation}
    \mathcal{L}_{ST} = \omega_s \mathcal{L}_{csc} + \omega_t \mathcal{L}_{ta},
\end{equation}
where $\omega_s$ and $\omega_t$ are learnable weights. The BH procedure controls FDR through a step-up process that adapts its rejection threshold based on the distribution of observed p-values. Unlike traditional single hypothesis testing, the BH procedure maintains FDR control at level $\alpha_{bh}$ even under arbitrary dependencies between tests, making it particularly suitable for CP where agents' behaviors may be correlated due to shared environmental factors or coordinated attacks.
We formulate the hypothesis test $\mathcal{H} = \{\mathcal{H}_0, \mathcal{H}_1\}$, where $\mathcal{H}_0: \mathcal{L}_{ST} \sim P_{normal}$ represents the null hypothesis that the combined score follows the distribution of normal agents, and $\mathcal{H}_1: \mathcal{L}_{ST} \nsim P_{normal}$ represents the alternative hypothesis.
Since this distribution is typically intractable, we compute conformal p-values using the calibration set $\mathcal{S}$:
\begin{equation}
    \hat{p} = \frac{1 + \left| \{s \in \mathcal{S} : s \geq \mathcal{L}_{ST}\} \right|}{1 + |\mathcal{S}|},
\end{equation}
\textcolor{black}{where $\vert \cdot \vert$ represents the cardinality of the set.} Following the BH procedure, we sort the p-values in ascending order $\hat{p}_{(1)} \leq \hat{p}_{(2)} \leq ... \leq \hat{p}_{(m)}$ for $m$ hypothesis tests. Let $j$ be the largest index satisfying:
\begin{equation}
    \hat{p}_{(j)} \leq \frac{j}{\textcolor{black}{m}}\alpha_{bh},
\end{equation}
Then, agent $i$ is classified as malicious when its p-value is less than or equal to $\hat{p}_{(j)}$, where $\alpha_{bh}$ controls the desired false detection rate.

\begin{algorithm}[t]
    \small
    \caption{Chain-based BEV Flow Matching \textcolor{black}{(BFM)}}
    \label{alg:GCP}
    \textbf{Input}: $\mathbf{O}_{e,i}^{t_{k}}$ (low-confidence detections), $\{\mathbf{Y}_{e,i}^{t_{k-j}}\}_{j=1}^K$ (cached detections)
    
    \textbf{Output}: $\mathbf{I}_{c}^{t_{k}}$, $\mathbf{I}_{u}^{t_{k}}$ (candidate and unmatched BEV flows)
    \begin{algorithmic}[1] 
    \Procedure{\texttt{BFM}}{$\mathbf{O}_{e,i}^{t_{k}}$}
        \State $\mathbf{I}_{c}^{t_{k}}, \mathbf{I}_{u}^{t_{k}} \leftarrow \{\}, \{\}$
        \Comment{\texttt{BEV Flow Initialization}}
        \If{$\mathbf{O}_{e,i}^{t_{k}} = \emptyset$} \Return $\mathbf{I}_{c}^{t_{k}}, \mathbf{I}_{u}^{t_{k}}$ \EndIf
        \For{$o_j \in \mathbf{O}_{e,i}^{t_{k}}$}
            \State $i_j \leftarrow [o_j]$, $curr \leftarrow o_j$, $matched \leftarrow \textbf{True}$
            \For{$k \leftarrow K-1$ \textbf{down to} $0$} 
            \Comment{\texttt{Chain Match}}
                \State $b_{max} \leftarrow \mathop{\arg\max}\limits_{b \in \mathbf{Y}_{e,i}^{t_{k}}} \operatorname{IoU}(b, curr)$
                \State $cost \leftarrow \operatorname{ReLU}(p_{curr} - p_{b}) + \phi(1-\operatorname{IoU}(b, curr))$
                \If{$cost < \tau$}
                    \State $i_j \leftarrow i_j \oplus b_{max}$, $curr \leftarrow b_{max}$
                \Else
                    \State $matched \leftarrow \textbf{False}$
                    \State \textbf{break}
                \EndIf
            \EndFor
           \If{$matched$}
           \Comment{\texttt{BEV Flow Concatenation}}
                \State $\mathbf{I}_{c}^{t_{k}} \leftarrow \mathbf{I}_{c}^{t_{k}} \cup \{i_j\}$
            \Else
                \State $\mathbf{I}_{u}^{t_{k}} \leftarrow \mathbf{I}_{u}^{t_{k}} \cup \{i_j\}$
            \EndIf
        \EndFor
        \State \Return $\mathbf{I}_{c}^{t_{k}}, \mathbf{I}_{u}^{t_{k}}$
    \EndProcedure
    \end{algorithmic}
\end{algorithm}

%% file: sections/experiment.tex
\section{Experiments}

\subsection{Experimental Setup}
\label{settings}

\textbf{Datasets.}
We conduct experiments using two data sets, namely, V2X-Sim dataset \citep{li2022v2x} and V2X-Flow dataset. V2X-Sim dataset is used for training and evaluating different CP models and CP defense methods, while V2X-Flow dataset is used for pre-training LSTM-AE in our \texttt{GCP}.

\ding{182} \textit{V2X-Sim Dataset.}
V2X-Sim \citep{li2022v2x} is a simulated dataset generated by the CARLA simulator \citep{Dosovitskiy17}. The dataset contains 10,000 frames of synchronized multi-view data captured from 6 different connected agents (5 vehicles and 1 RSU), including LiDAR point clouds (32-beam, 70m range) and RGB images, along with 501,000 annotated 3D bounding boxes containing object attributes.
The data is split into training, validation, and test sets with a ratio of 8:1:1.

\ding{183} \textit{V2X-Flow Dataset.} To facilite the pre-training of LSTM-AE used in our \texttt{GCP}, we construct a V2X-Flow dataset based on V2X-Sim. We separate the annotations of the V2X-Sim dataset to generate the BEV flow of each bounding box using the chain-based BEV flow matching method mentioned in Section \ref{flow}.


\noindent \textbf{Attack Settings.} \textcolor{blue}{To simulate realistic stealthy attacks, we model temporal patterns where malicious agents attack intermittently. Given $N_a$ total agents, $m$ malicious agents, and time horizon $T$, we define the attack ratio $\lambda$ as the proportion of malicious messages in the total traffic. We evaluate three modes where the total malicious load $\lambda N_aT$ is distributed among $m$ attackers via a truncated normal distribution $\mathcal{N}(\frac{\lambda N_aT}{m}, \sigma^2)$, reflecting varying attack capabilities:}

\ding{182} \textit{Random attack process (R-mode)}  \citep{Li_2023_ICCV, zhao2024maliciousagentdetectionrobust}. 
This mode follows a uniform random distribution in which malicious messages are randomly distributed over the time period $T$, with $\lambda N_aT$ total malicious messages from $m$ malicious agents, simulating unpredictable attack patterns. 

\ding{183} \textit{Poisson attack process (P-mode)}  \citep{5696753}. This mode models bursty attack patterns (e.g., DDoS attacks) in which intense activities occur in short durations. At each time step $t \in T$, the number of malicious messages $N_t$ follows a Poisson distribution: \begin{equation}
    P\left(N_t = k\right) = \frac{(\lambda N_a)^{k}exp\left(-\lambda N_a\right)}{k!},
\end{equation}
where $k$ denotes the number of malicious messages at time step $t$, and $\lambda N_a$ represents the mean rate. 

\ding{184} \textit{Susceptible-Infectious process (S-mode)}  \citep{9998460}. This mode simulates progressive attack scenarios where malicious behaviors spread through the network over time, similar to virus propagation. The evolution of the number of malicious messages $N_t$ follows:
\begin{equation}
    \frac{dN_t}{dt} = \gamma N_t(1 - \frac{N_t}{\lambda N_aT}),
\end{equation}
For adversarial perturbation generation, we evaluate our method against three representative white-box attacks: Projected Gradient Descent (PGD) \citep{madry2018towards}, Carlini \& Wagner (C\&W) \citep{7958570}, and Basic Iterative Method (BIM) (Please refer to Appendix \ref{appendix:attacks} for implementation details). Moreover, we introduce our proposed BAC attack to exploit the vulnerabilities in CP systems specifically. To maintain real-time attack capability, the BAC mask generation adopts a slow update strategy without requiring per-frame updates. In our experiments, we set the mask update rate to 0.5 FPS.

\noindent \textbf{Implementation Details.}
Our \texttt{GCP} is implemented with PyTorch. Each agent's locally captured LiDAR or camera data is first encoded into a BEV feature map with size $16 \times 16$ and 512 channels.
The encoder network adopts a ResNet-style architecture with 5 layers of feature encoding, with output feature dimensions progressively increasing from 32 to 512 channels while spatial dimensions decrease from 256 to 16.
The decoder consists of three convolutional blocks with channel dimensions progressively decreasing from 512 to 64.
The detection head follows a two-branch design (classification and regression) with standard convolutional layers, batch normalization, and ReLU activation. It uses an anchor-based detection decoder \citep{8578474} with multiple anchor sizes per location.
The fusion method is V2VNet \citep{10.1007/978-3-030-58536-5_36}.
The LSTM-AE model consists of 2 LSTM layers with hidden dimension 32. We train the model using Adam optimizer (lr=0.001) with the default history sequence length 5.
\textcolor{blue}{All experiments are conducted on a server with 2 Intel(R) Xeon(R) Silver 4410Y CPUs, a single NVIDIA RTX A5000 GPU, and 512 GB RAM.}

\noindent \textbf{Baselines and Evaluation Metrics.}
We compare our method with state-of-the-art defenses ROBOSAC \citep{Li_2023_ICCV}, MADE \citep{zhao2024maliciousagentdetectionrobust}, and \textcolor{blue}{CP-Guard \citep{hucpguard2025}}. \textcolor{blue}{We also introduce a simple Gated Late Fusion as a baseline, which aggregates proposal boxes from all connected agents and filters them using a consistency voting mechanism.} Reference settings include Upper-bound (clean CP), Lower-bound (ego-only), and No Defense. Evaluation metrics include Average Precision (AP@0.5, AP@0.7) for accuracy and Frames Per Second (FPS) for efficiency.

\subsection{Quantitative Results}

\noindent \textcolor{blue}{\textbf{Benchmark Comparison.} Table \ref{tab:quantitative_results_v2xsim} evaluates \texttt{GCP} on V2X-Sim. First, while No Defense suffers severe degradation, all defenses show effectiveness. Notably, CP-Guard and Gated Late Fusion are competitive against conventional attacks. However, despite MADE slightly outperforming ROBOSAC, all baselines are vulnerable to our BAC attack.}
\textcolor{blue}{In contrast, \texttt{GCP} excels. For conventional attacks, it achieves consistent improvements (0.47\%-0.79\% AP@0.5). The advantage is pronounced under BAC attack, where \texttt{GCP} surpasses CP-Guard, Gated Late Fusion, MADE, and ROBOSAC by 9.42\%, 6.26\%, 12.64\%, and 8.53\% respectively, maintaining near-upper-bound performance.}
\textcolor{blue}{Across R/P/S-Modes, \texttt{GCP} demonstrates consistent robustness where baselines degrade, validating our spatial-temporal defense against diverse attacks.}

\begin{table*}[t]
    \caption{\textcolor{blue}{\textbf{Comparative results under different attack methods and modes.} Settings: $m=2, \lambda=0.25, \Delta_i=\Delta_o=0.5$, iterations=10. \textbf{Bold}/\underline{Underlined}: Best/Second-best (excl. upper-bound). $\downarrow$: Significant drop.}}
    \label{tab:quantitative_results_v2xsim}
    \centering 
    \renewcommand{\arraystretch}{0.65}
    \setlength{\tabcolsep}{8pt}
    \resizebox{1\linewidth}{!}{
    \scriptsize
    \begin{tabular}{l|c|cc|cc|cc|cc}
        \hline 
        \multirow{2}{*}{\textbf{Method}} & \multirow{2}{*}{\textbf{Mode}} & \multicolumn{2}{c|}{PGD attack} & \multicolumn{2}{c|}{C\&W attack} & \multicolumn{2}{c|}{BIM attack} & \multicolumn{2}{c}{BAC attack}\\
        & & AP@0.5 & AP@0.7  & AP@0.5 & AP@0.7 & AP@0.5 & AP@0.7  & AP@0.5 & AP@0.7 \\
        \hline 
        Upper-bound & --- & 80.52 & 78.65 & 80.52 & 78.65 & 80.52 & 78.65 & 80.52 & 78.65 \\ 
        \hline
        \multirow{4}{*}{\textbf{\texttt{GCP} (Ours)}} & Avg. & \textbf{77.54} & \textbf{76.51} & \textbf{76.81} & \textbf{75.56} & \textbf{77.40} & \textbf{76.46} & \textbf{76.64} & \textbf{75.64} \\
        & R & 76.88 & 75.78 & 76.90 & 75.16 & 76.72 & 75.74 & 75.80 & 74.96 \\
        & P & 77.37 & 76.34 & 77.13 & 76.20 & 77.38 & 76.43 & 76.29 & 75.09 \\
        & S & 78.36 & 77.41 & 76.41 & 75.32 & 78.11 & 77.22 & 77.82 & 76.87 \\
        \hline
        \multirow{4}{*}{\textcolor{blue}{CP-Guard}} & \textcolor{blue}{Avg.} & \textcolor{blue}{76.54} & \textcolor{blue}{75.24} & \textcolor{blue}{75.82} & \textcolor{blue}{74.45} & \textcolor{blue}{76.12} & \textcolor{blue}{74.68} & \textcolor{blue}{67.22}$\downarrow$ & \textcolor{blue}{63.88}$\downarrow$ \\
        & \textcolor{blue}{R} & \textcolor{blue}{76.12} & \textcolor{blue}{74.98} & \textcolor{blue}{75.45} & \textcolor{blue}{73.88} & \textcolor{blue}{76.35} & \textcolor{blue}{74.88} & \textcolor{blue}{67.78} & \textcolor{blue}{63.95} \\
        & \textcolor{blue}{P} & \textcolor{blue}{76.25} & \textcolor{blue}{74.55} & \textcolor{blue}{75.92} & \textcolor{blue}{74.38} & \textcolor{blue}{76.45} & \textcolor{blue}{75.38} & \textcolor{blue}{66.88} & \textcolor{blue}{64.55} \\
        & \textcolor{blue}{S} & \textcolor{blue}{77.25} & \textcolor{blue}{76.18} & \textcolor{blue}{76.08} & \textcolor{blue}{75.08} & \textcolor{blue}{75.55} & \textcolor{blue}{73.78} & \textcolor{blue}{66.98} & \textcolor{blue}{63.15} \\
        \hline
        \multirow{4}{*}{\textcolor{blue}{Gated Late Fusion}} & \textcolor{blue}{Avg.} & \textcolor{blue}{71.18} & \textcolor{blue}{68.82} & \textcolor{blue}{70.52} & \textcolor{blue}{67.98} & \textcolor{blue}{71.08} & \textcolor{blue}{68.75} & \underline{\textcolor{blue}{70.38}} & \underline{\textcolor{blue}{68.02}} \\
        & \textcolor{blue}{R} & \textcolor{blue}{70.65} & \textcolor{blue}{68.12} & \textcolor{blue}{70.58} & \textcolor{blue}{67.58} & \textcolor{blue}{70.42} & \textcolor{blue}{68.08} & \textcolor{blue}{69.62} & \textcolor{blue}{67.42} \\
        & \textcolor{blue}{P} & \textcolor{blue}{71.02} & \textcolor{blue}{68.64} & \textcolor{blue}{70.78} & \textcolor{blue}{68.48} & \textcolor{blue}{71.05} & \textcolor{blue}{68.72} & \textcolor{blue}{70.05} & \textcolor{blue}{67.52} \\
        & \textcolor{blue}{S} & \textcolor{blue}{71.95} & \textcolor{blue}{69.62} & \textcolor{blue}{70.18} & \textcolor{blue}{67.72} & \textcolor{blue}{71.68} & \textcolor{blue}{69.42} & \textcolor{blue}{71.48} & \textcolor{blue}{69.12} \\
        \hline
        \multirow{4}{*}{MADE} & Avg. & \underline{77.07} & \underline{75.75} & \underline{76.28} & \underline{74.87} & \underline{76.61} & \underline{75.06} & 64.00$\downarrow$ & 54.48$\downarrow$ \\
        & R & 76.67 & 75.56 & 75.81 & 74.28 & 76.82 & 75.28 & 65.63 & 56.92 \\
        & P & 76.76 & 75.04 & 76.29 & 74.81 & 76.96 & 75.88 & 65.11 & 55.95 \\
        & S & 77.79 & 76.66 & 76.39 & 75.42 & 76.06 & 74.01 & 61.28 & 50.58 \\
        \hline
        \multirow{4}{*}{ROBOSAC} & Avg. & 73.31 & 71.54 & 73.90 & 72.16 & 73.48 & 71.70 & 68.11$\downarrow$ & 64.77$\downarrow$ \\
        & R & 74.64 & 72.93 & 74.52 & 72.78 & 74.67 & 73.19 & 68.68 & 64.82 \\
        & P & 73.98 & 72.84 & 73.63 & 72.45 & 73.63 & 72.07 & 67.25 & 65.48 \\
        & S & 71.31 & 68.84 & 73.54 & 71.26 & 72.14 & 69.85 & 68.40 & 64.02 \\
        \hline
        \multirow{4}{*}{No Defense} & Avg. & 36.65$\downarrow$ & 35.92$\downarrow$ & 17.70$\downarrow$ & 15.14$\downarrow$ & 38.27$\downarrow$ & 37.28$\downarrow$ & 54.83$\downarrow$ & 45.11$\downarrow$ \\
        & R & 36.50 & 35.64 & 18.03 & 15.52 & 37.07 & 36.18 & 55.19 & 43.99 \\
        & P & 39.01 & 38.35 & 18.92 & 16.23 & 36.59 & 35.47 & 55.03 & 45.07 \\
        & S & 34.45 & 33.78 & 16.14 & 13.66 & 41.16 & 40.19 & 54.26 & 46.28 \\
        \hline
        Lower-bound & --- & 64.08 & 61.99 & 64.08 & 61.99 & 64.08 & 61.99 & 64.08 & 61.99 \\
        \hline
    \end{tabular}
    }
    \vspace{-3mm}
\end{table*}

\begin{table}[htbp]
    \centering
    \caption{\textcolor{blue}{Performance comparison (AP@0.5 / AP@0.7) under Independent vs. Colluding attacks ($m=2$).}}
    \label{tab:collusion}
    \resizebox{1\linewidth}{!}{
    \textcolor{blue}{
    \begin{tabular}{l|l|cc|cc}
    \hline
    \multirow{2}{*}{\textbf{Attack}} & \multirow{2}{*}{\textbf{Method}} & \multicolumn{2}{c|}{\textbf{Independent}} & \multicolumn{2}{c}{\textbf{Colluding}} \\
    \cline{3-6}
     & & \textbf{AP@0.5} & \textbf{AP@0.7} & \textbf{AP@0.5} & \textbf{AP@0.7} \\
    \hline
    \multirow{2}{*}{PGD} & No Defense & 36.65 & 35.92 & 34.35 & 33.66 \\
     & \texttt{GCP} (Ours) & 77.54 & 76.51 & 74.51 & 73.53 \\
    \hline
    \multirow{2}{*}{C\&W} & No Defense & 17.70 & 15.14 & 16.42 & 14.05 \\
     & \texttt{GCP} (Ours) & 76.81 & 75.56 & 73.89 & 72.68 \\
    \hline
    \multirow{2}{*}{BIM} & No Defense & 38.27 & 37.28 & 35.88 & 34.91 \\
     & \texttt{GCP} (Ours) & 77.40 & 76.46 & 74.65 & 73.72 \\
    \hline
    \multirow{2}{*}{BAC} & No Defense & 54.83 & 45.11 & 51.92 & 42.85 \\
     & \texttt{GCP} (Ours) & 76.64 & 75.64 & 74.28 & 73.35 \\
    \hline
    \end{tabular}
    }
    }
\end{table}

\begin{table}[t]
    \centering
    \caption{\textcolor{blue}{Impact of the number of malicious agents ($m$) on detection performance (under PGD attack).}}
    \label{tab:scalability}
    \resizebox{1\linewidth}{!}{
    \textcolor{blue}{
    \begin{tabular}{c|cc|cc}
    \hline
    \multirow{2}{*}{\textbf{\# Mal. ($m$)}} & \multicolumn{2}{c|}{\textbf{AP@0.5}} & \multicolumn{2}{c}{\textbf{AP@0.7}} \\
    \cline{2-5}
     & \textbf{No Defense} & \textbf{\texttt{GCP} (Ours)} & \textbf{No Defense} & \textbf{\texttt{GCP} (Ours)} \\
    \hline
    0 (Benign) & 80.52 & 78.42 & 78.65 & 77.51 \\
    1 & 50.32 & 78.32 & 48.42 & 77.05 \\
    2 (Default) & 36.65 & 77.54 & 35.92 & 76.51 \\
    3 & 22.11 & 73.08 & 21.56 & 71.88 \\
    \hline
    \end{tabular}
    }
    }
\end{table}

\begin{table}[t]
    \caption{\textcolor{blue}{\textbf{Ablation study.} \texttt{GCP} vs. variants (\texttt{GCP-S/T}). Settings: $m=2, \lambda=0.25, \Delta_i=\Delta_o=0.5$.}}
    \label{tab:ablation}
    \centering 
    
    \resizebox{1\linewidth}{!}{
    \small
    \renewcommand{\arraystretch}{0.95}
    \setlength{\tabcolsep}{12pt}
    \begin{tabular}{l|cc|cc}
        \hline 
        \multirow{2}{*}{\textbf{Method}} & \multicolumn{2}{c|}{PGD attack} & \multicolumn{2}{c}{BAC attack}\\
        & AP@0.5 & AP@0.7  & AP@0.5 & AP@0.7 \\
        \hline 
        Upper-bound & 80.52 & 78.65 & 80.52 & 78.65  \\ 
        \hline
        \textbf{\texttt{GCP} (Ours)} & \textbf{77.54} & \textbf{76.51} & \textbf{76.64} & \textbf{75.64}  \\ 
        \texttt{GCP-S} & 77.21 & 76.10 & 69.23 & 62.38  \\ 
        \texttt{GCP-T} & 70.16 & 67.33 & 66.01 & 58.79  \\ 
        \hline
        No Defense & 36.65 & 35.92 & 54.83 & 45.11 \\ 
        \hline 
    \end{tabular}
    }
    \vspace{-3mm}
\end{table}

\begin{figure*}[t]
    \centering
    \includegraphics[width=0.95\linewidth]{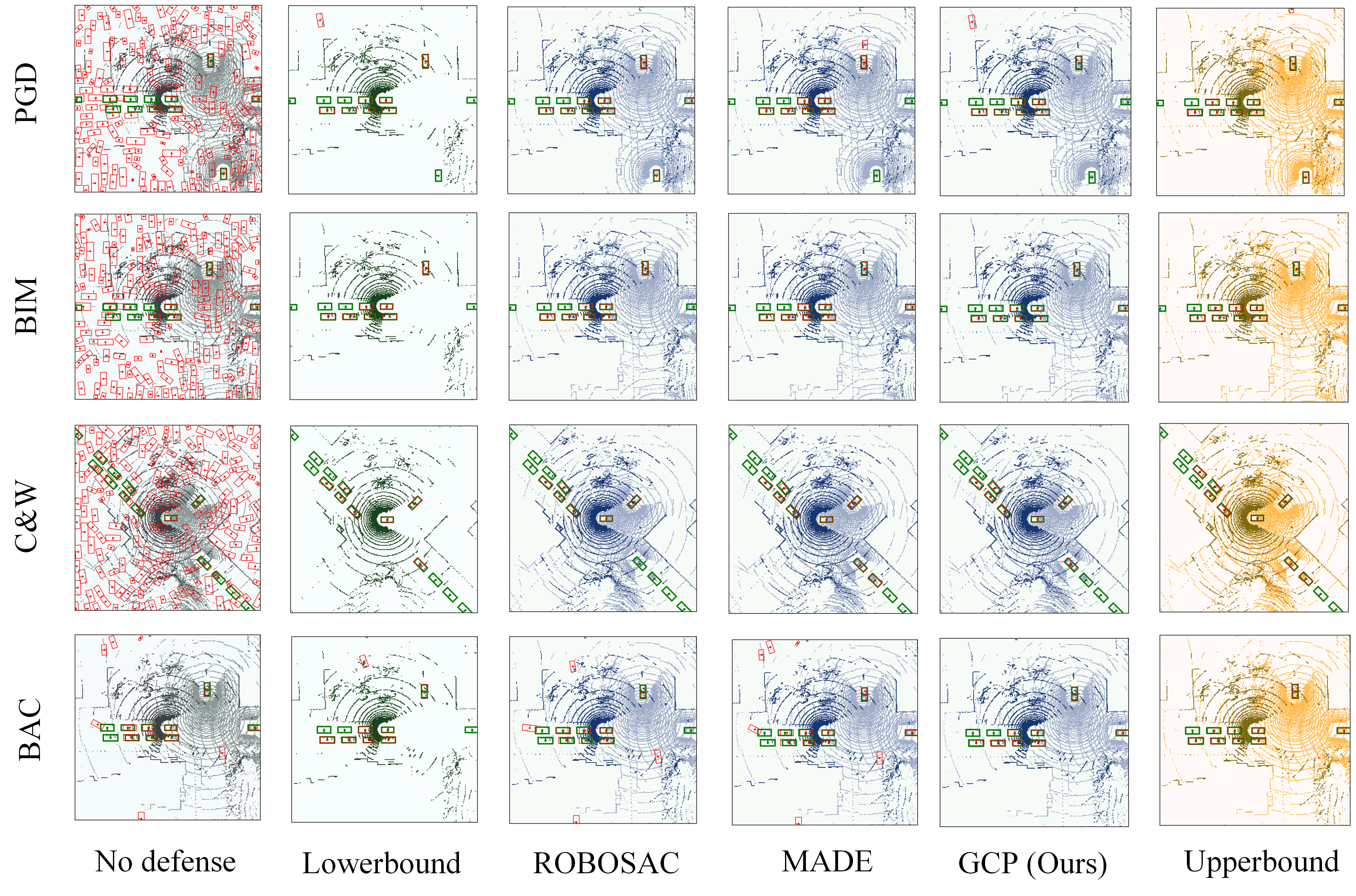}
    \caption{\textbf{Visualization of 3D detection results} on V2X-Sim dataset. Attack settings: number of malicious agents = 2; attack ratio = 0.25; input/output perturbation budget = 0.5. Scene ID: 8, Frame ID: 81, 65, 30, 47 (from top to bottom). \textcolor{red}{Red} boxes are predictions while the \textcolor{ForestGreen}{green} ones are GT.}
    \label{fig:visualization_detection}
    \vspace{-3mm} 
\end{figure*}

\noindent \textcolor{blue}{\textbf{Robustness under Different Attack Configurations.} We first evaluate stability under varying intensities and attacker counts. Table \ref{tab:quantitative_results_pert} shows that under intense attacks ($m = 3$, $\lambda = 0.5$), \texttt{GCP} maintains stability with only 5.6\% degradation, significantly outperforming baselines. It also shows consistent performance ($<$0.5\% variation) across different perturbation budgets. Furthermore, Table \ref{tab:scalability} confirms resilience to increasing attacker counts. In benign scenarios ($m=0$), \texttt{GCP} matches the Upper-bound, ensuring minimal impact. As $m$ increases, \texttt{GCP} maintains robust performance (around 77\% for $m \le 2$) while No Defense degrades catastrophically. This confirms \texttt{GCP}'s reliability in highly compromised networks.}

\noindent \textcolor{blue}{\textbf{Robustness against More Attack Strategies.} We further evaluate \texttt{GCP} against two advanced strategies. \ding{182} \textit{Coordinated Attacks}: As shown in Table \ref{tab:collusion}, even when agents optimize a joint adversarial objective, \texttt{GCP} remains highly effective (AP@0.5 $>$ 73\%). \ding{183} \textit{LiDAR Spoofing Attacks}: We implement a physically realistic Ray-Casting attack \cite{294490} that injects synthesized points by calculating intersections between LiDAR rays and ghost vehicle meshes. Although this generates plausible raw data perturbations, the resulting ghost boxes are often sparse and intermittent due to optimization constraints. \texttt{GCP} effectively identifies these anomalies, achieving an 84.5\% defense success rate (at $K=5$).}

\begin{table*}[t]
    \caption{\textcolor{blue}{\textbf{Performance under different attack intensities.} Settings: Moderate ($m=2, \lambda=0.25$) vs. Intense ($m=3, \lambda=0.5$). \textbf{Bold}/\underline{Underlined}: Best/Second-best (excl. upper-bound). $\downarrow$: Significant drop.}}
    \label{tab:quantitative_results_pert}
    \centering 
    \renewcommand{\arraystretch}{0.90}
    \setlength{\tabcolsep}{8pt}
    \resizebox{1\linewidth}{!}{
        \scriptsize
        \begin{tabular}{l|c|cc|cc|cc|cc}
            \hline 
            \multirow{2}{*}{\textbf{Method}} & \multirow{2}{*}{\textbf{BAC Attack}} & \multicolumn{2}{c|}{$\Delta_{i} = 0.5, \Delta_{o} = 0.5$} & \multicolumn{2}{c|}{$\Delta_{i} = 0.5, \Delta_{o} = 0.1$} & \multicolumn{2}{c|}{$\Delta_{i} = 1.0, \Delta_{o} = 0.5$} & \multicolumn{2}{c}{$\Delta_{i} = 1.0, \Delta_{o} = 0.1$}\\
            & & AP@0.5 & AP@0.7  & AP@0.5 & AP@0.7 & AP@0.5 & AP@0.7  & AP@0.5 & AP@0.7 \\
            \hline 
            Upper-bound & --- & 80.52 & 78.65 & 80.52 & 78.65 & 80.52 & 78.65 & 80.52 & 78.65 \\ 
            \hline
            \multirow{2}{*}{\textbf{\texttt{GCP} (Ours)}} & Moderate & \textbf{75.80} & \textbf{74.96} & \textbf{75.23} & \textbf{74.02} & \textbf{75.58} & \textbf{74.28} & \textbf{75.51} & \textbf{74.69} \\
            & Intense & \textbf{70.22} & \textbf{68.57} & \textbf{70.03} & \textbf{68.36} & \textbf{70.07} & \textbf{68.59} & \textbf{69.77} & \textbf{68.20} \\
            \hline
            \multirow{2}{*}{\textcolor{blue}{CP-Guard}} & \textcolor{blue}{Moderate} & \underline{\textcolor{blue}{72.15}} & \underline{\textcolor{blue}{68.25}} & \underline{\textcolor{blue}{69.88}} & \underline{\textcolor{blue}{65.55}} & \underline{\textcolor{blue}{67.55}} & \underline{\textcolor{blue}{61.25}} & \underline{\textcolor{blue}{69.25}} & \underline{\textcolor{blue}{64.55}} \\
            & \textcolor{blue}{Intense} & \textcolor{blue}{63.55} & \textcolor{blue}{57.88} & \underline{\textcolor{blue}{64.55}} & \textcolor{blue}{58.25} & \textcolor{blue}{62.15} & \textcolor{blue}{54.25} & \underline{\textcolor{blue}{64.15}} & \textcolor{blue}{56.95} \\
            \hline
            \multirow{2}{*}{\textcolor{blue}{Gated Late Fusion}} & \textcolor{blue}{Moderate} & \textcolor{blue}{69.62} & \textcolor{blue}{67.42} & \textcolor{blue}{69.12} & \textcolor{blue}{66.58} & \textcolor{blue}{69.42} & \textcolor{blue}{66.82} & \textcolor{blue}{69.35} & \textcolor{blue}{67.18} \\
            & \textcolor{blue}{Intense} & \underline{\textcolor{blue}{64.48}} & \underline{\textcolor{blue}{61.65}} & \textcolor{blue}{64.32} & \underline{\textcolor{blue}{61.48}} & \underline{\textcolor{blue}{64.35}} & \underline{\textcolor{blue}{61.62}} & \textcolor{blue}{64.08} & \underline{\textcolor{blue}{61.32}} \\
            \hline
            \multirow{2}{*}{MADE} & Moderate & 65.63 & 56.92 & 65.80 & 57.26 & 65.39 & 55.56 & 66.93 & 56.79 \\
            & Intense & 51.94$\downarrow$ & 40.87$\downarrow$ & 54.22$\downarrow$ & 41.38$\downarrow$ & 53.21$\downarrow$ & 40.61$\downarrow$ & 51.04$\downarrow$ & 36.25$\downarrow$ \\
            \hline
            \multirow{2}{*}{ROBOSAC} & Moderate & 68.68 & 64.82 & 66.26 & 62.38 & 64.12 & 58.08 & 65.64 & 61.42 \\
            & Intense & 60.09 & 54.88 & 61.20 & 55.07 & 59.08 & 51.14 & 61.02 & 53.87 \\
            \hline
            \multirow{2}{*}{No Defense} & Moderate & 55.19$\downarrow$ & 43.99$\downarrow$ & 65.05 & 55.44 & 55.92$\downarrow$ & 42.91$\downarrow$ & 59.54 & 47.27$\downarrow$ \\
            & Intense & 35.63$\downarrow$ & 25.35$\downarrow$ & 47.82$\downarrow$ & 32.55$\downarrow$ & 33.40$\downarrow$ & 17.98$\downarrow$ & 41.85$\downarrow$ & 23.24$\downarrow$ \\
            \hline
        \end{tabular}
        }
        \vspace{-3mm}
\end{table*}

\begin{figure*}[t]
    \centering
    \includegraphics[width=0.95\linewidth]{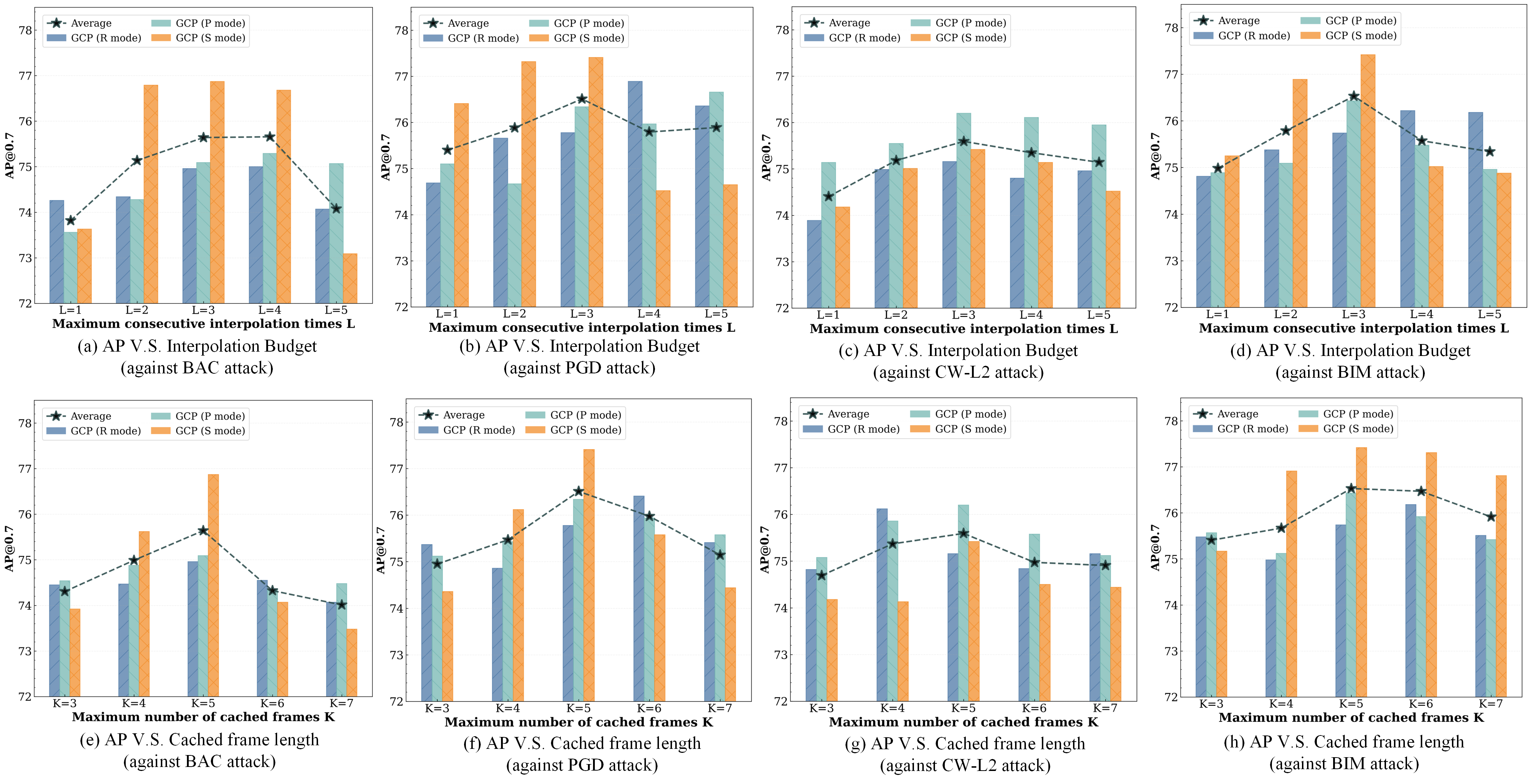}
    \caption{\textbf{Comparative results under different cached frame length and consecutive KF interpolation times}. Attack settings: $m = 2$, $\lambda = 0.25$; $\Delta_i = \Delta_o = 0.5$. (a)-(d): AP@0.7 under different attacks and interpolation budgets; (e)-(h): AP@0.7 under different attacks and cached frame length.}
    \label{fig:cached_frames}
    \vspace{-3mm} 
\end{figure*}

\begin{table}[htbp]
    \centering
    \caption{\textcolor{blue}{Resource consumption and runtime breakdown.}}
    \label{tab:efficiency}
    \resizebox{1\linewidth}{!}{
    \textcolor{blue}{
    \begin{tabular}{l|c|c|c}
    \hline
    \multicolumn{4}{c}{\textbf{Resource Consumption}} \\
    \hline
    Metric & No Defense & \texttt{GCP} (Ours) & Overhead \\
    \hline
    GPU Memory (MB) & 28,020 & 28,138 & +118 \\
    GPU Utilization (\%) & 0.6 & 0.9 & +0.3 \\
    CPU Utilization (\%) & 10.4 & 6.7 & -3.7 \\
    \hline
    \multicolumn{4}{c}{\textbf{Runtime Breakdown}} \\
    \hline
    Module & \multicolumn{2}{c|}{Time (ms)} & Equiv. FPS \\
    \hline
    Spatial Defense (CSCLoss) & \multicolumn{2}{c|}{3.2} & - \\
    Temporal Defense (Flow Gen.) & \multicolumn{2}{c|}{1.8} & - \\
    Temporal Defense (LSTM-AE) & \multicolumn{2}{c|}{21.0} & - \\
    \hline
    \textbf{Total Defense Latency} & \multicolumn{2}{c|}{\textbf{26.0}} & \textbf{38.6} \\
    \hline
    \end{tabular}
    }
    }
\end{table}

\begin{table}[t]
    \caption{\textcolor{blue}{\textbf{Attack and defense speed (FPS).} Settings: $m = 2, \lambda = 0.25, \Delta_i = \Delta_o = 0.5$.}}
    \label{tab:attack_latency}
    \centering 
    
    \resizebox{1\linewidth}{!}{
    \small
    \renewcommand{\arraystretch}{0.85}
    \setlength{\tabcolsep}{12pt}
    \begin{tabular}{l|c|ccc}
        \hline 
        \multirow{2}{*}{\textbf{Attack Method}} & \textbf{Attack} & \multicolumn{3}{c}{\textbf{Defense Speed (FPS)}} \\
        & \textbf{Speed (FPS)} & MADE  & ROBOSAC  & \texttt{GCP} \\
        \hline
        PGD attack & 25.3 & 55.8 & 27.1 & 38.6 \\
        C\&W attack & 18.2 & 38.5 & 21.6 & 30.2 \\
        BIM attack & 24.8 & 54.6 & 28.7 & 37.5 \\
        BAC attack & 22.4 & 82.8 & 44.6 & 47.6 \\
        \hline
    \end{tabular}
    }
    \vspace{-3mm}
\end{table}

\noindent \textcolor{blue}{\textbf{Overhead Analysis.} We analyze the computational efficiency on a single NVIDIA RTX A5000 GPU using the V2X-Sim test set under the default attack setting ($m=2, \lambda=0.25, \Delta_i=\Delta_o=0.5$). As detailed in Table \ref{tab:efficiency}, the average total defense latency per frame is around 26.0 ms, which is dominated by the temporal LSTM-AE inference. This latency corresponds to a processing speed of around 38.6 FPS, as compared with other methods in Table \ref{tab:attack_latency}. Since this speed significantly exceeds the typical 10-20 Hz sensor frequency (e.g., LiDAR), \texttt{GCP} can verify collaborative messages within the generation cycle. This ensures a 0-frame detection delay (where attacks at frame $t$ are identified immediately), preventing perception bottlenecks. Furthermore, \texttt{GCP} is resource-efficient, incurring zero bandwidth overhead and negligible GPU memory cost (+118 MB), making it ideal for resource-constrained deployment. For extreme constraints, further optimizations like model distillation or FP16 quantization could be applied.}

\noindent \textcolor{blue}{\textbf{Generalizability to Different CP Model.} We extend the evaluation to include DiscoNet \citep{NEURIPS2021_f702defb} and When2com \citep{Liu_2020_CVPR} besides the default V2VNet. DiscoNet utilizes distilled collaboration while When2com employing handshake communication for collaboration. Table \ref{tab:generalizability} (averaged over three attack modes) shows that \texttt{GCP} consistently protects these diverse CP models, demonstrating its broad applicability and robustness across different collaborative fusion frameworks.}

\subsection{Ablation Studies}

\noindent \textcolor{blue}{\textbf{Ablation on Defense Components.}} We evaluate the contributions of our defense components. First, we compare \texttt{GCP} with its spatial-only (\texttt{GCP-S}) and temporal-only (\texttt{GCP-T}) variants. As shown in Table \ref{tab:ablation}, both components contribute to the overall defense. For PGD attacks, the spatial component is dominant, with \texttt{GCP-S} achieving 77.21\% AP@0.5, comparable to the full model. However, for the challenging BAC attack, neither component alone suffices, with significant drops in \texttt{GCP-S} (7.41\%) and \texttt{GCP-T} (10.63\%). This confirms the necessity of our joint spatial-temporal design. \textcolor{black}{Notably, \texttt{GCP-S} alone outperforms ROBOSAC \citep{Li_2023_ICCV} and MADE \citep{zhao2024maliciousagentdetectionrobust}, validating our CSCLoss design.}

\noindent \textcolor{blue}{\textbf{Ablation on Hyper-Parameters.}} We analyze the frame cache length ($K$) and consecutive KF interpolation times ($L$). As shown in Fig. \ref{fig:cached_frames}, moderate values yield optimal performance. The cache size peaks around $K=5$ (AP@0.5 of 76.87\% under BAC S-mode), balancing context and noise. For interpolation times, performance is optimal at $L=3$ (77.42\% AP@0.5 for BIM), as excessive interpolation may introduce cumulative errors.

\begin{figure*}[t]
    \centering
    \includegraphics[width=0.95\linewidth]{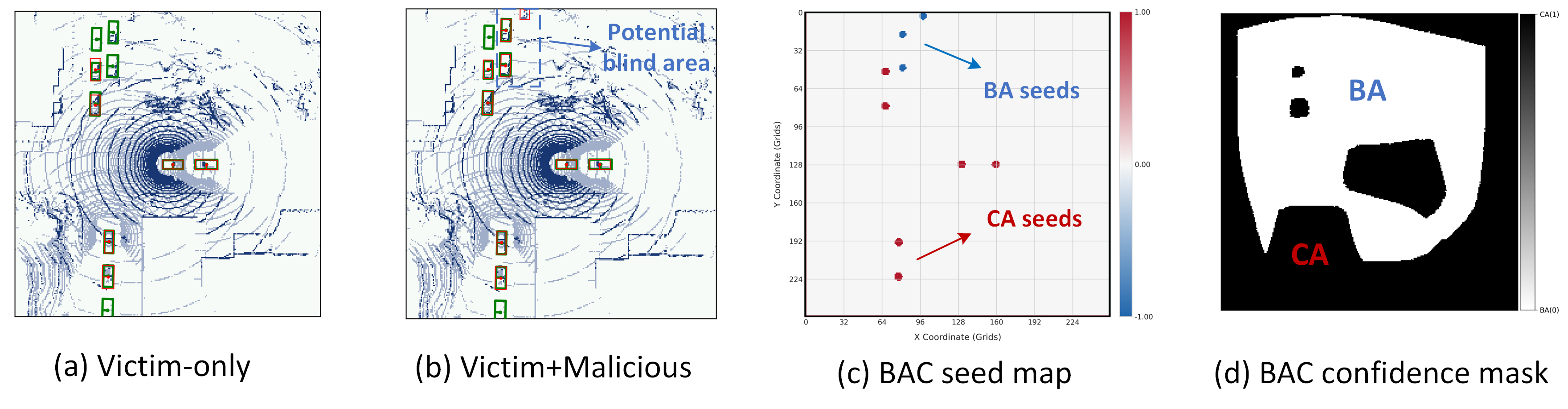}
    \caption{\textbf{Visualization of the BAC attack pipeline} on V2X-Sim dataset. (a) Detection results using only victim vehicle's local perception; (b) Enhanced detection results through CP; (c) Initial BAC seed map generated from differential detection results; (d) Refined BAC confidence mask obtained through blind region segmentation. \textcolor{red}{Red} boxes are predictions while the \textcolor{ForestGreen}{green} ones are GT. }
    \label{fig:BAC_attack}
    \vspace{-3mm} 
\end{figure*}

\begin{figure}[t]
    \centering
    \includegraphics[width=0.95\linewidth]{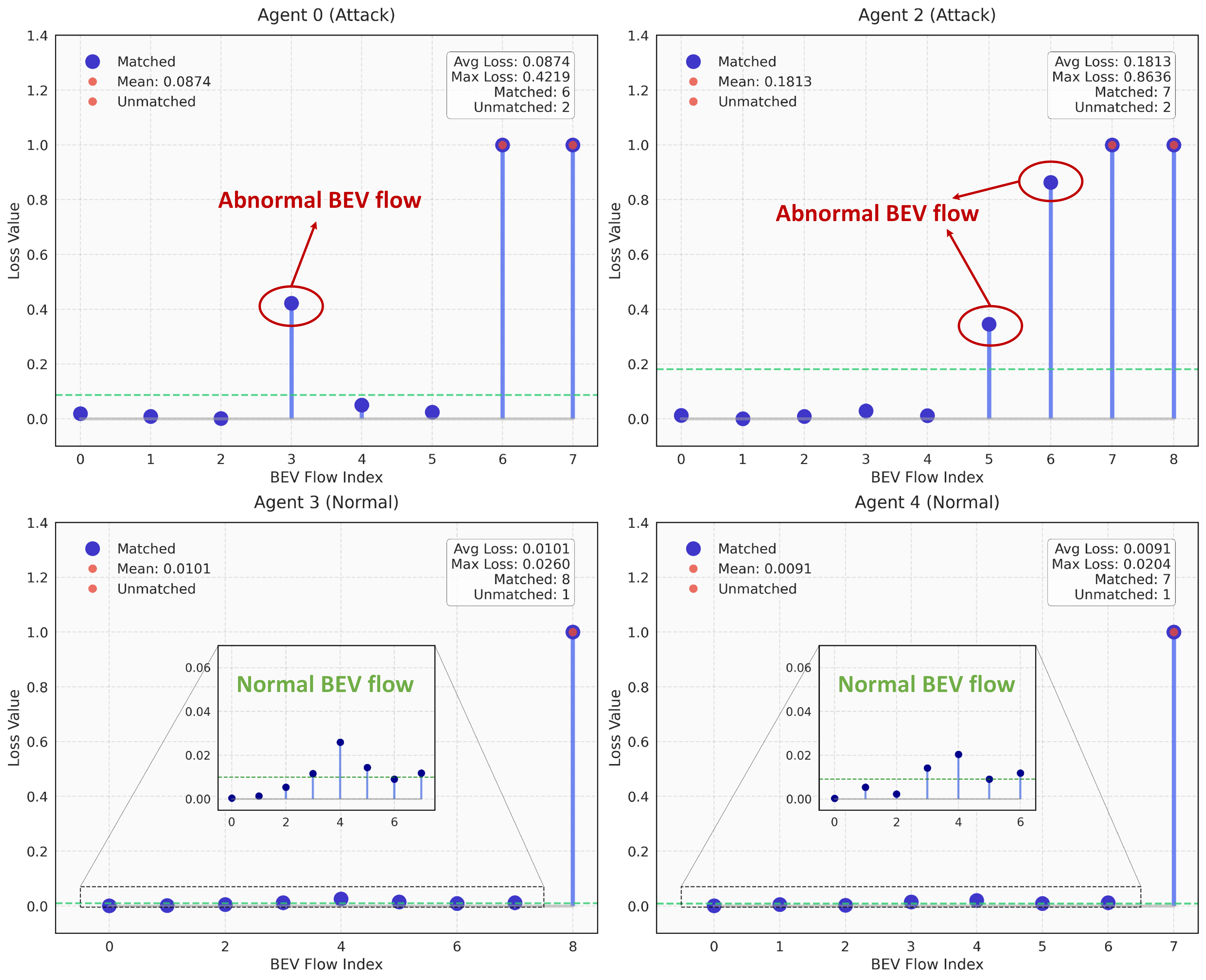}
    \caption{\textbf{BEV flow reconstruction loss distribution.} Agent 0 and 2 are under BAC attack ($\Delta_o = 0.5$) while agent 3 and 4 are normal. Scene ID: 8, Frame ID: 61.}
    \label{fig:flow_loss}
    \vspace{-3mm} 
\end{figure}

\begin{figure}[t]
    \vspace{-3mm}
    \centering
    \includegraphics[width=0.95\linewidth]{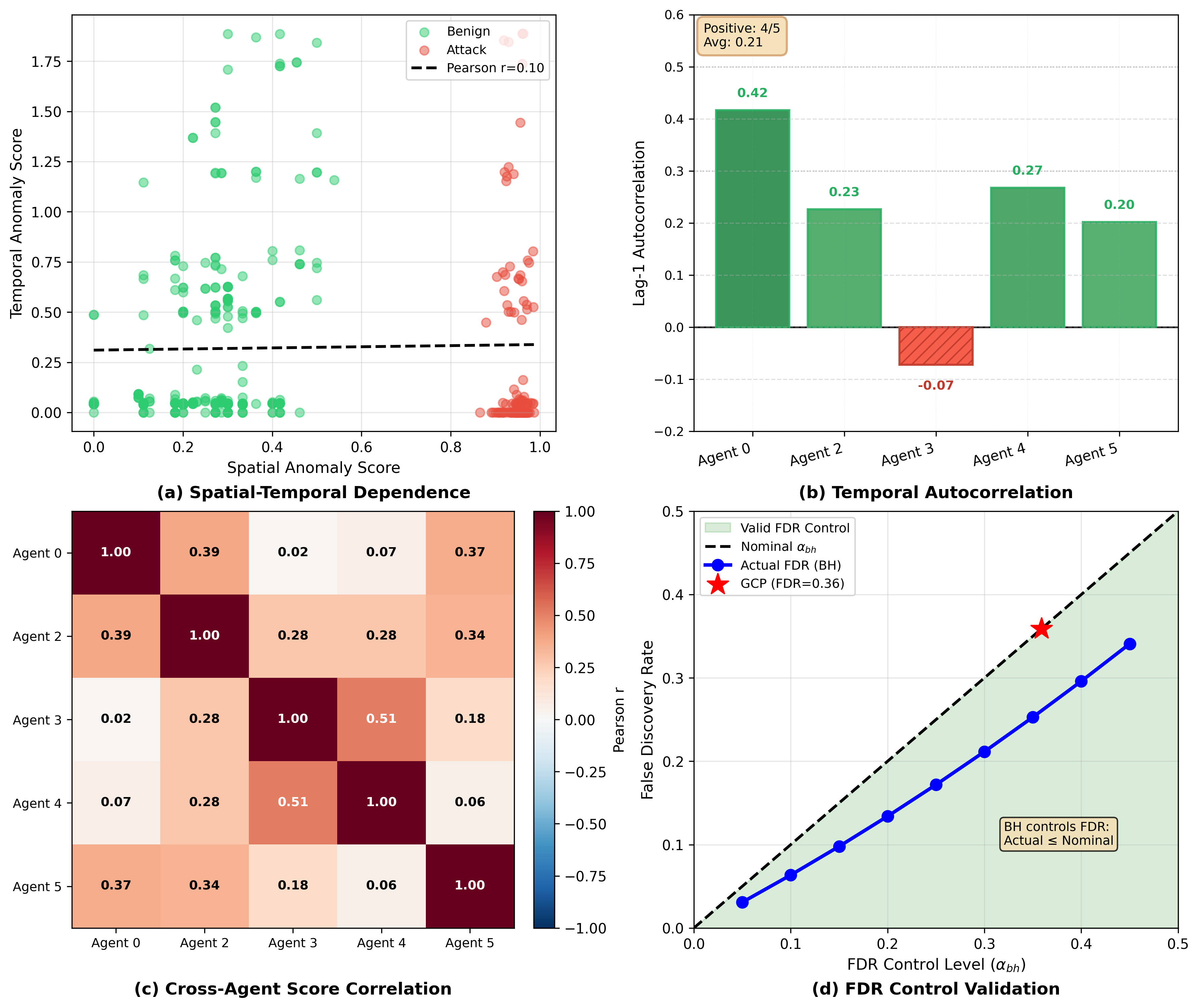}
    \caption{\textcolor{blue}{\textbf{Statistical dependency analysis and FDR control validation.} It validates the PRDS assumption via positive correlations and demonstrates the effectiveness of the chosen operating point.}}
    \label{fig:bh_analysis}
    \vspace{-5mm} 
\end{figure}

\begin{table}[t]
    \caption{\textcolor{blue}{Generalizability of GCP across different CP models.}}
    \label{tab:generalizability}
    \centering
    \resizebox{1\linewidth}{!}{
    \textcolor{blue}{
    \begin{tabular}{l|c|cc|cc}
    \hline
    \multirow{2}{*}{\textbf{CP Model}} & \multirow{2}{*}{\textbf{Method}} & \multicolumn{2}{c|}{AP@0.5} & \multicolumn{2}{c}{AP@0.7} \\
    \cline{3-6}
     & & No Def. & \texttt{GCP} & No Def. & \texttt{GCP} \\
    \hline
    \multirow{4}{*}{DiscoNet} & PGD & 37.05 & \textbf{78.35} & 36.25 & \textbf{77.28} \\
     & C\&W & 17.88 & \textbf{77.62} & 15.28 & \textbf{76.35} \\
     & BIM & 38.69 & \textbf{78.21} & 37.62 & \textbf{77.25} \\
     & BAC & 55.42 & \textbf{77.45} & 45.58 & \textbf{76.42} \\
    \hline
    \multirow{4}{*}{When2com} & PGD & 23.65 & \textbf{49.65} & 22.05 & \textbf{46.85} \\
     & C\&W & 11.45 & \textbf{49.18} & 9.32 & \textbf{46.28} \\
     & BIM & 24.68 & \textbf{49.52} & 22.88 & \textbf{46.82} \\
     & BAC & 35.32 & \textbf{49.08} & 27.65 & \textbf{46.35} \\
    \hline
    \end{tabular}
    }
    }
    \vspace{-3mm}
\end{table}

\subsection{Qualitative Results}

\noindent \textcolor{blue}{\textbf{Visualization of Attack Pipeline.}} Fig. \ref{fig:BAC_attack} illustrates the complete pipeline of the BAC attack on the V2X-Sim dataset. The malicious agent analyzes detection results from the victim's local perception (Fig. \ref{fig:BAC_attack}(a)) to identify blind spots. By comparing with CP results (Fig. \ref{fig:BAC_attack}(b)), it pinpoints regions where the victim relies heavily on collaborative messages. This differential analysis yields an initial seed map (Fig. \ref{fig:BAC_attack}(c)), which is refined into a BAC confidence mask (Fig. \ref{fig:BAC_attack}(d)) via blind region segmentation to guide targeted perturbations.

\noindent \textcolor{blue}{\textbf{Visualization of Defense Performance.}} We visualize the collaborative 3D detection performance in Fig. \ref{fig:visualization_detection}. While conventional attacks (PGD, C\&W, BIM) generate obvious perturbations that are easily detected, our BAC attack optimizes output-space perturbations in blind regions, demonstrating superior stealthiness and causing significant degradation in baselines like ROBOSAC and MADE. In contrast, \texttt{GCP} maintains robust detection through spatial-temporal consistency verification, achieving results comparable to the upper bound.

\noindent \textcolor{blue}{\textbf{Visualization of BEV Flow Reconstruction Loss.}} Fig. \ref{fig:flow_loss} analyzes the BEV flow reconstruction loss distribution. Normal agents (Agent 3 and 4) show consistently low losses (mean $\approx$ 0.01) and stable temporal consistency. In contrast, agents under BAC attack (Agent 0 and 2) exhibit significantly higher maximum losses ($>$ 0.4) and more unmatched flows. These high-loss outliers and temporal inconsistencies serve as reliable indicators for detecting stealthy attacks.

\noindent \textcolor{blue}{\textbf{Visualization of Statistical Validity.} We validate the statistical assumptions on a small calibration set comprising 389 agent-frame samples from V2X-Sim Scene 8. The BH procedure requires the test statistics to satisfy Positive Regression Dependence on a Subset (PRDS). As shown in Fig. \ref{fig:bh_analysis}, we observe consistently non-negative cross-agent correlations ($r \in [0.02, 0.51]$) due to shared field-of-views, alongside positive temporal autocorrelation from physical continuity. This positive dependence structure satisfies the PRDS condition, ensuring the validity of FDR control. Fig. \ref{fig:bh_analysis}(d) further confirms that our chosen operating point ($\alpha_{bh}=0.36$) achieves 100\% attack recall while maintaining conservative FDR control, effectively balancing safety and benign performance.}

\subsection{Limitation and Future Work}

\noindent \textcolor{blue}{Currently, we focus on defending against untargeted adversarial attacks without temporal optimization. While sophisticated ``slow-drift'' attacks could potentially evade detection by generating temporally consistent long-term trajectories, such methods are currently unexplored in the context of real-time multi-agent systems due to the optimization difficulities. Therefore, investigating the feasibility of such stealthy threats and developing corresponding defense mechanisms remains an important direction for future research.}

%% file: sections/conclusion.tex
\section{Conclusion}

In this paper, we have devised a novel blind area confusion (BAC) attack to show that collaborative perception (CP) systems are vulnerable to malicious attacks even with existing outlier-based CP defense mechanisms. The key innovation of the BAC attack lies in the blind region segmentation-based local perturbation optimization. To counter such attacks, we have proposed our \texttt{GCP}, a robust CP defense framework utilizing spatial and temporal contextual information through confidence-scaled spatial concordance loss and LSTM-AE-based temporal BEV flow reconstruction. These components are integrated via a spatial-temporal Benjamini-Hochberg test to generate reliable anomaly scores for malicious agent detection. 
\textcolor{blue}{Extensive experiments demonstrate \texttt{GCP}'s superior robustness against diverse attacks while preserving benign performance. These results validate the efficacy of spatial-temporal analysis in guarding CP systems. We believe our work establishes a solid foundation for secure multi-agent collaboration in future autonomous driving applications.}

\newpage

%% file: sections/bio.tex
\vspace{-10pt}
\newpage
\begin{IEEEbiography}[{\includegraphics[width=1in,height=1.25in,clip,keepaspectratio]{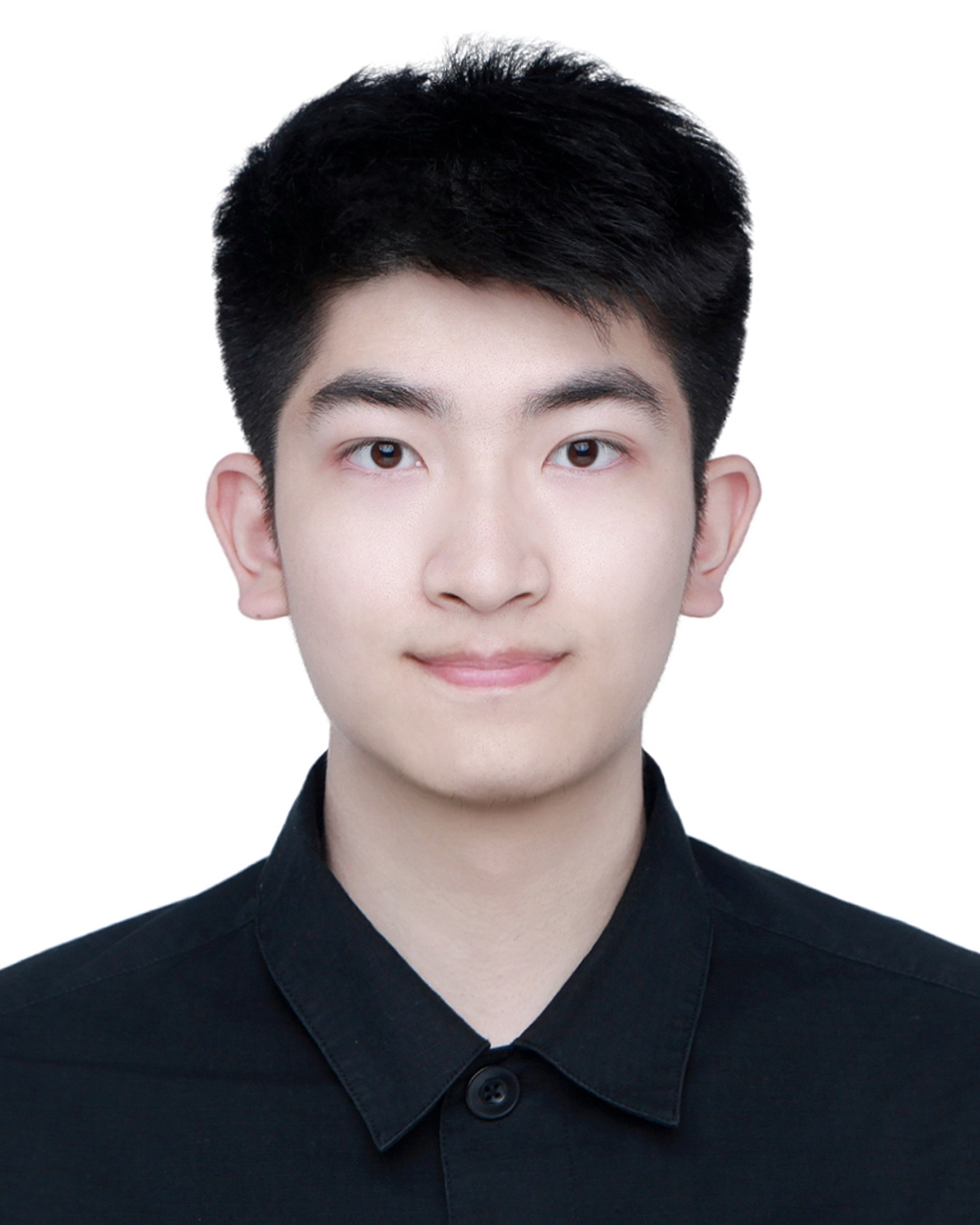}}]{Yihang Tao}
       received the B.S. degree from the School of Information Science and Engineering, Southeast University, Nanjing, China, in 2021 and received his M.S. degree from the School of Computer Science, Shanghai Jiao Tong University, Shanghai, China, in 2024. Currently, he is pursuing his PhD degree in the Department of Computer Science at City University of Hong Kong. His current research interests include world model, spatial intelligence, and autonomous driving. He serves as the PC member for CVPR, ICML, ICLR, etc.
\end{IEEEbiography}
\vspace{-20pt}

\begin{IEEEbiography}[{\includegraphics[width=1in,height=1.25in,clip,keepaspectratio]{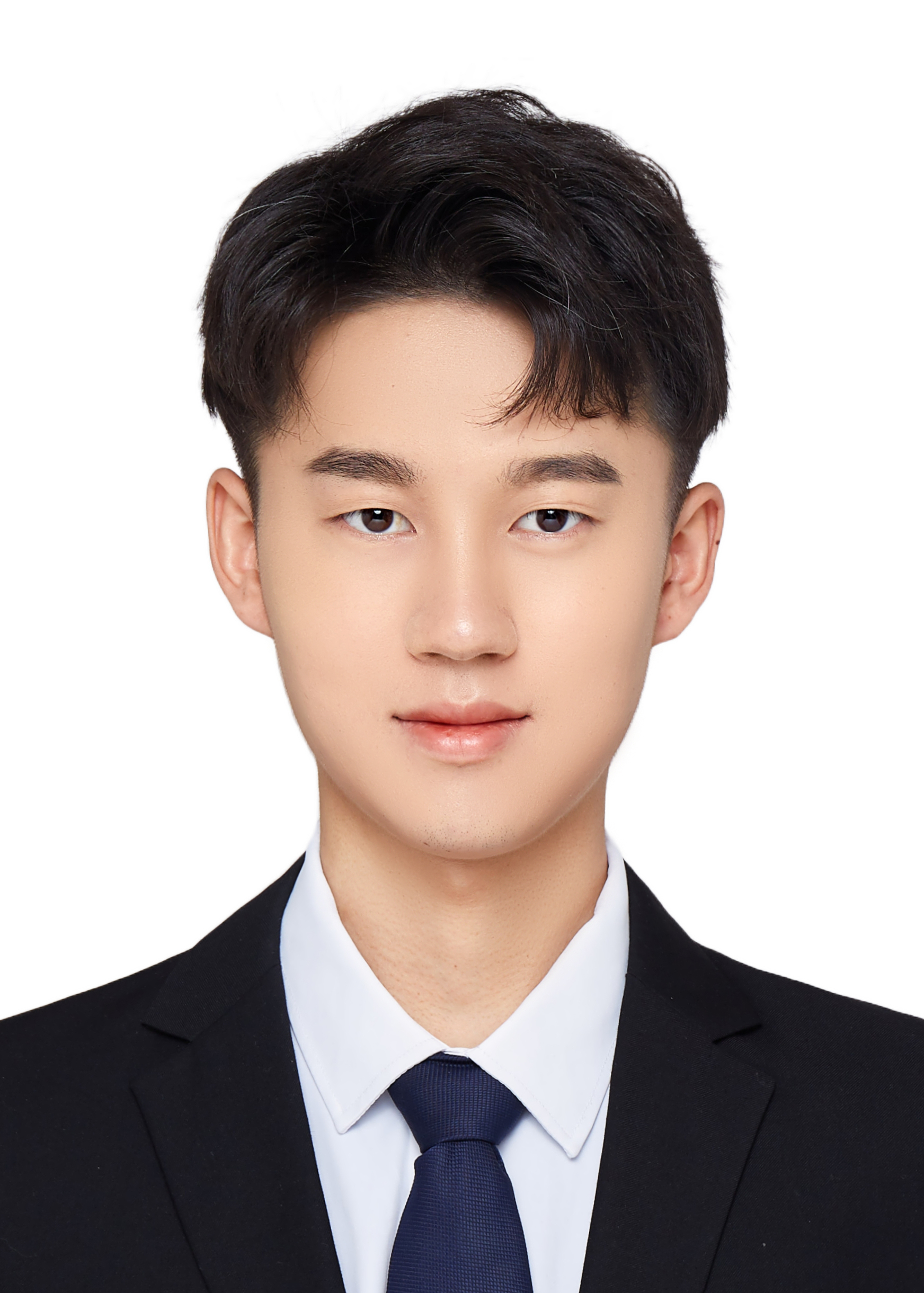}}]{Senkang Hu}
    received his B.S. degree in electronic and information engineering from Beijing Institute of Technology, Beijing, China, in 2022. He is pursuing his PhD in the Department of Computer Science at City University of Hong Kong, Hong Kong. His research interests include connected and autonomous driving, vehicle-to-vehicle collaborative perception, and LLMs.
    \end{IEEEbiography}

\begin{IEEEbiography}[{\includegraphics[width=1in,height=1.25in,clip,keepaspectratio]{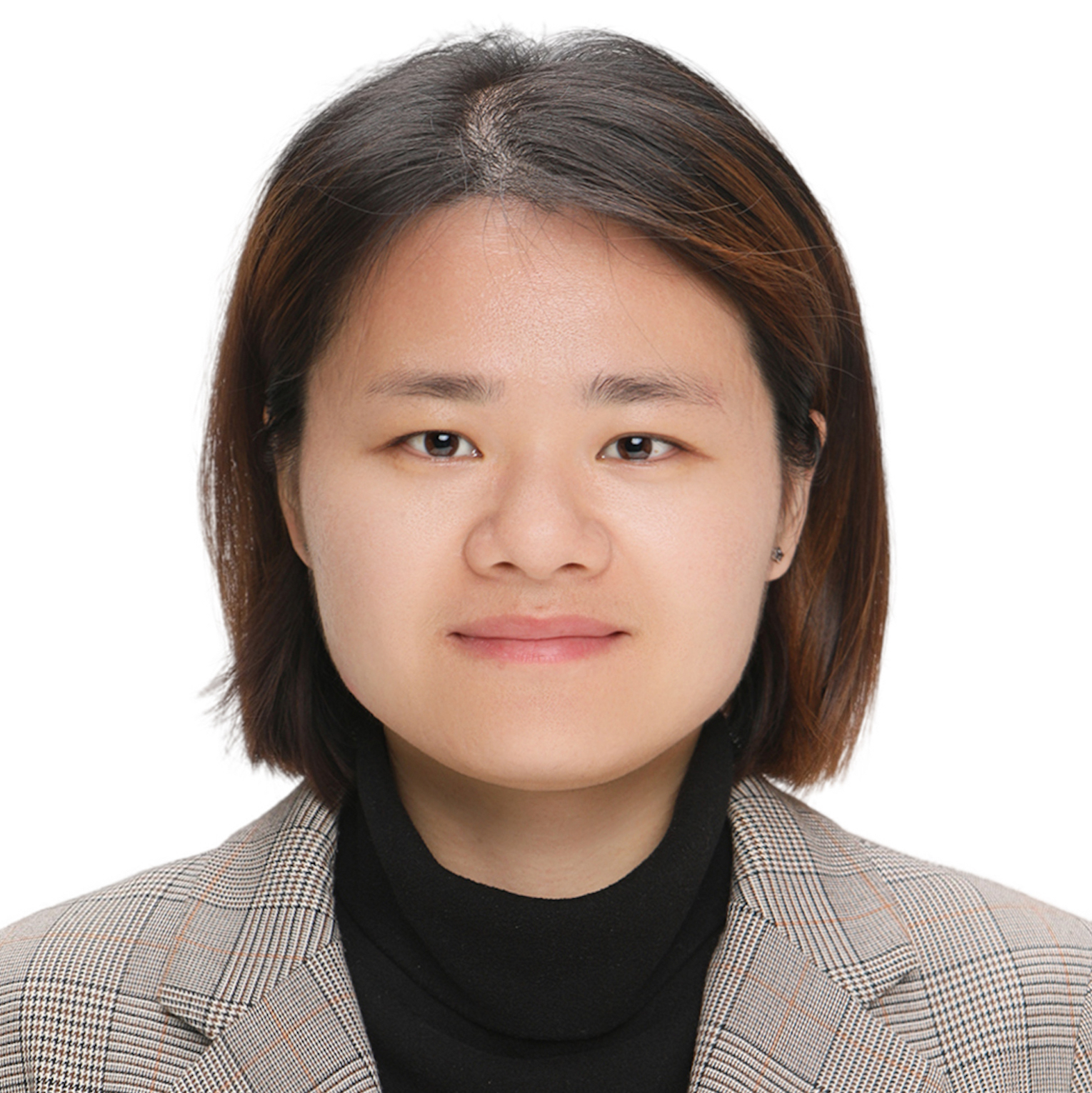}}]{Yue Hu} is currently a postdoctoral researcher in the Department of Robotics, University of Michigan, Ann Arbor, USA. She received her PhD from the Cooperative Medianet Innovation Center at Shanghai Jiao Tong University in 2024. She received her MS and BE degrees in information engineering from Shanghai Jiao Tong University, Shanghai, China, in 2020 and 2017, respectively. Her research interests include multi-agent collaboration, communication efficiency, and 3D vision.
    \end{IEEEbiography}
    \vspace{-20pt}

\begin{IEEEbiography}[{\includegraphics[width=1in,height=1.25in,clip,keepaspectratio]{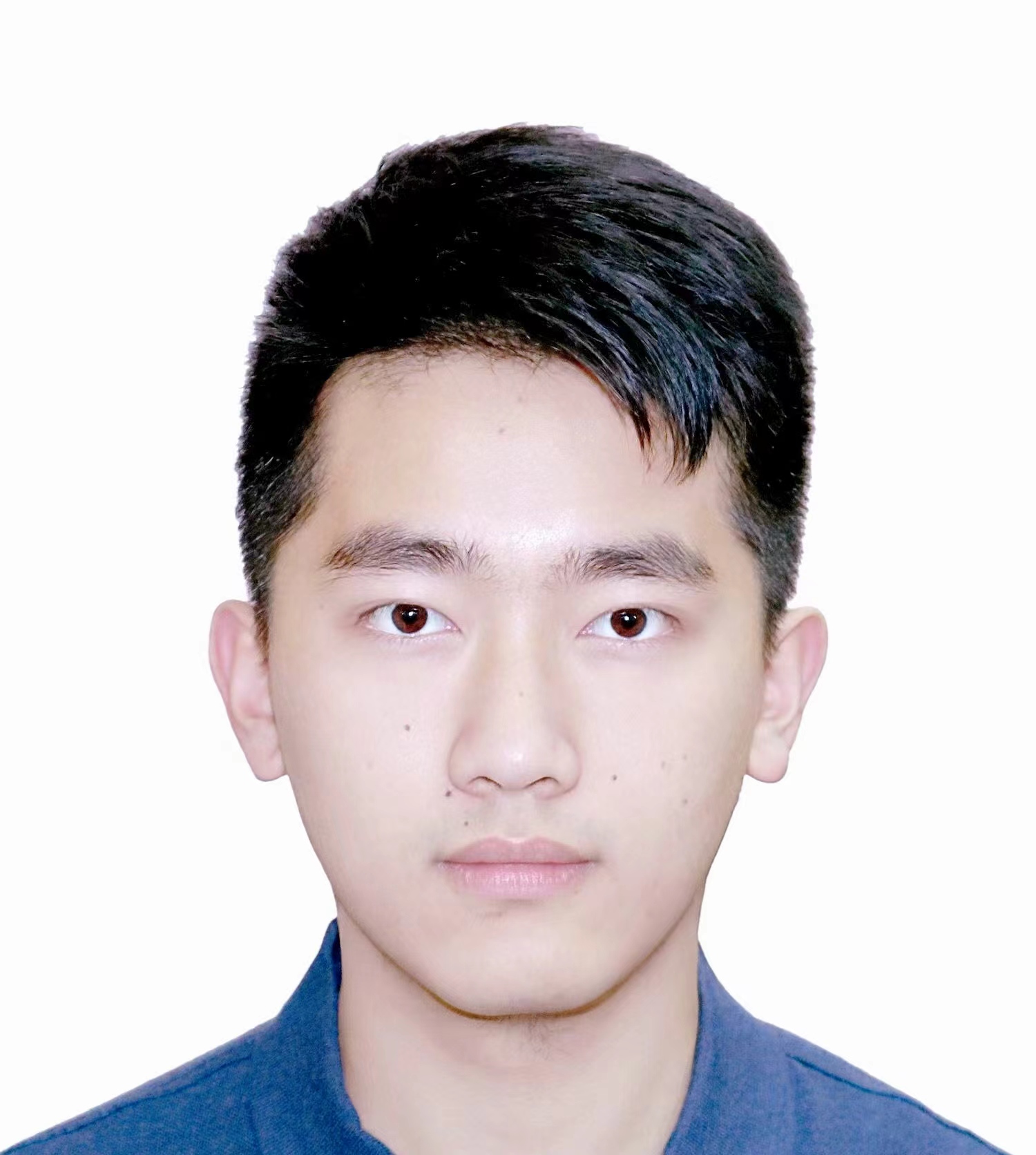}}]{Haonan An}
    received his BE degree in Telecommunication Engineering from Huazhong University of Science and Technology, Wuhan, China, in 2023, and his MS in Signal Processing from the School of Electrical and Electronic Engineering, Nanyang Technological University, Singapore, in 2024. He is pursuing his Ph.D. at the City University of Hong Kong. His research interests include digital watermarking, AI security, and autonomous driving.
    \end{IEEEbiography}
 \vspace{-20pt}

\begin{IEEEbiography}[{\includegraphics[width=1in,height=1.25in,clip,keepaspectratio]{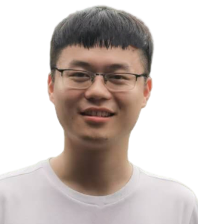}}]{Hangcheng Cao} is currently a postdoctoral fellow in the Department of Computer Science, City University of Hong Kong. He obtained the Ph.D. degree in the College of Computer Science and Electronic Engineer, Hunan University, China, in 2023. He studied as a joint PhD student in the School of Computer Science and Engineering, Nanyang Technological University, Singapore, in 2022. He has published papers in IEEE S\&P, ACM Ubicomp/IMWUT, IEEE INFOCOM, IEEE ICDCS, IEEE TMC, ACM ToSN, IEEE Communications Magazine, Information Fusion, etc. His research interests lie in the area of IoT security.  
    \end{IEEEbiography}
 \vspace{-20pt}

\begin{IEEEbiography}[{\includegraphics[width=1in,height=1.25in,clip,keepaspectratio]{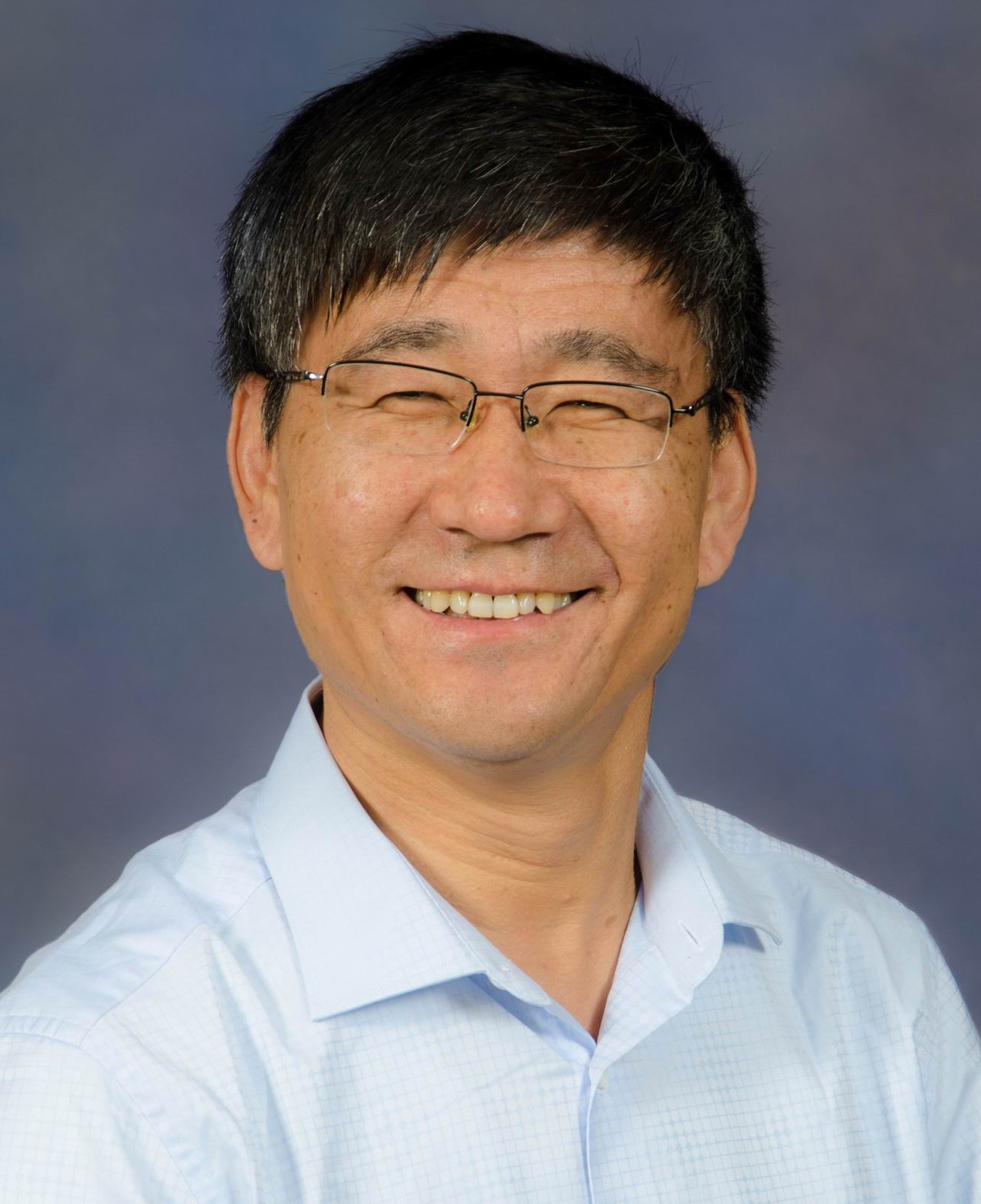}}]{Yuguang Fang}
(S’92, M’97, SM’99, F’08) received his MS from Qufu Normal University, China
in 1987, his PhD from Case Western Reserve University, Cleveland, Ohio, USA, in 1994, and another PhD from Boston University, Boston, Massachusetts, USA, in 1997. He joined the Department of Electrical and Computer Engineering at University of Florida in 2000 as an assistant professor, then was promoted to associate professor in 2003, full professor in 2005, and distinguished professor in 2019, respectively. Since August 2022, he has been a Global STEM Scholar and Chair Professor with the Department of Computer Science at City University of Hong Kong.

Prof. Fang received many awards, including the US NSF CAREER Award
(2001), US ONR Young Investigator Award (2002), 2018 IEEE Vehicular Technology Outstanding Service Award, IEEE Communications Society's
AHSN Technical Achievement Award (2019), CISTC Technical Recognition
Award (2015) and WTC Recognition Award (2014), and 2010-2011 UF
Doctoral Dissertation Advisor/Mentoring Award. He held multiple professorships, including the Changjiang Scholar Chair Professorship (2008-2011),
Tsinghua University Guest Chair Professorship (2009-2012), University of
Florida Foundation Preeminence Term Professorship (2019-2022), and University of Florida Research Foundation Professorship (2017-2020, 2006-2009). He served as the Editor-in-Chief of IEEE Transactions on Vehicular
Technology (2013-2017) and IEEE Wireless Communications (2009-2012)
and serves/served on several editorial boards of journals, including Proceedings
of the IEEE (2018-2023), ACM Computing Surveys (2017-present), ACM
Transactions on Cyber-Physical Systems (2020-present), IEEE Transactions
on Mobile Computing (2003-2008, 2011-2016, 2019-present), IEEE Transactions on Communications (2000-2011), and IEEE Transactions on Wireless
Communications (2002-2009). He served as the Technical Program Co-Chair of IEEE INFOCOM’2014. He is a Member-at-Large of the Board of
Governors of IEEE Communications Society (2022-2024) and the Director of
Magazines of IEEE Communications Society (2018-2019). He is a fellow of
ACM, IEEE, and AAAS.
\end{IEEEbiography}

            

%% file: sections/appendix.tex
\newpage

\section{KF-based BEV Flow Interpolation}
\label{appendix:Kalman}

Given the state transition equations for intermittent BEV flow, we can directly apply these to the Kalman filter (KF) framework for both prediction and state update, and thereby interpolate the missing values. Assume that system state and observation noises are additive white Gaussian noises and that the state transition and observation models are linear, 
we first define the state vector and the state transition matrix of BEV flow:
\begin{equation}
    o_j = [x_j^1,y_j^1, \cdots, x_j^4, y_j^4]^\top \in \mathbb{R}^8,
\end{equation}
where $o_j$ represents the BEV flow state vector, containing 4 corner points of a bounding box in the BEV detection map, with $(x_j^k, y_j^k)$ being the coordinates of the $k$-th corner point. Then, the state transition equation can be represented as:
\begin{equation}
    o_{j,k+1} = \mathbf{F}_k o_{j,k} + \mathbf{w}_k,
\end{equation}
where $\mathbf{F}_k$ is the state transition matrix constructed based on the equations provided, assuming no external control inputs besides the physical model-based linear relationships:
\begin{equation}
    \mathbf{F}_k = \begin{bmatrix}
    \mathbf{I}_{4} & \Delta t \mathbf{I}_{4} \\
    \mathbf{0}_{4} & \mathbf{I}_{4}
    \end{bmatrix},
\end{equation}
where $\mathbf{I}_{4}$ represents $4\times 4$ identity matrix, $\mathbf{0}_{4}$ is $4\times 4$ zero matrix, and $\Delta t$ is the time difference between frame $k$ and frame $k-1$.
Based on the state transition matrix above, there are two stages for KF, namely, the prediction stage and the update stage. In the prediction stage, both the state and the error covariance are predicted, given as:
\begin{equation}
    \hat{o}_{j,k+1|k} = \mathbf{F}_k \hat{o}_{j,k|k},
\end{equation}
\begin{equation}
    \mathbf{P}_{k+1|k} = \mathbf{F}_k \mathbf{P}_{k|k} \mathbf{F}_k^T + \mathbf{Q}_k,
\end{equation}
where $\mathbf{Q}_k$ is the process noise covariance matrix, which needs to be set based on practical scenarios.
In the update stage, assuming the observation vector $\mathbf{z}_k$ directly reflects all state variables, we have the observation model as follows:
\begin{equation}
    \mathbf{z}_k = \mathbf{H}_k o_{j,k} + \mathbf{v}_k,
\end{equation}
where $\mathbf{H}_k$ is the observation matrix, which can be simplified to the identity matrix $\mathbf{I}_8$, if all state variables are directly observable.
The KF update process includes updating Kalman gain, state estimate, and error covariance as follows:
\begin{equation}
    \mathbf{K}_k = \mathbf{P}_{k+1|k} \mathbf{H}_k^T (\mathbf{H}_k \mathbf{P}_{k+1|k} \mathbf{H}_k^T + \mathbf{R}_k)^{-1},
\end{equation}
\begin{equation}
   \hat{o}_{j,k+1|k+1} = \hat{o}_{j,k+1|k} + \mathbf{K}_k (\mathbf{z}_k - \mathbf{H}_k \hat{o}_{j,k+1|k}),
\end{equation}
\begin{equation}
    \mathbf{P}_{k+1|k+1} = (\mathbf{I}_8 - \mathbf{K}_k \mathbf{H}_k) \mathbf{P}_{k+1|k},
\end{equation}
where $\mathbf{K}_k$ is the Kalman gain, $\hat{o}_{j,k+1|k+1}$ is the state estimate, $\mathbf{P}_{k+1|k+1}$ is the error covariance, and $\mathbf{R}_k$ is the observation noise covariance matrix.

\section{Implementation of Adversarial Attacks}
\label{appendix:attacks}
In this paper, we evaluate our proposed \texttt{GCP} and other baselines by implementing two adversarial attacks: 
\begin{itemize}
    \item \textbf{Projected Gradient Descent (PGD) Attack} \citep{madry2018towards}: PGD attack introduces a random initialization step to the adversarial example generation process. The mathematical expression for PGD is initiated by adding uniformly distributed noise to the original input:
    \begin{equation}
    \mathbf{F}_k^0 = \mathbf{F}_k + \text{Uniform}(-\Delta, \Delta),
    \end{equation}
    where $\Delta$ is a predefined perturbation limit.
    Subsequent iterations adjust the adversarial example by moving in the direction of the gradient of the loss function:
     \begin{equation}
            \mathbf{F}_k^{t+1} = \Pi_{\Delta}\{\mathbf{F}_k^t + \alpha \cdot \text{sign}(\nabla_{\mathbf{F}_k^t} \mathcal{L}(\mathbf{F}_k^t, \mathbf{y}))\},
    \end{equation}
    where $t$ denotes the iteration index, $\alpha$ is the step size and $\Pi{\Delta}$ represents the projection operation that confines the perturbation within the allowable range. This procedure is typically repeated for a predefined number of iterations, with settings such as $T=15$ and $\alpha=0.1$ often used.

    \item \textbf{Carini \& Wagner (C\&W) Attack} \citep{7958570}: The C\&W attack focuses on identifying the minimal perturbation $\delta$ that leads to a misclassification, formulated as the following optimization problem:
    \begin{equation}
    \min_{\delta} |\delta|_p + c \cdot f(\mathbf{F}_k + \delta),
    \end{equation}
    where the function $|\cdot|_p$ measures the size of the perturbation using the $L_p$ norm, while $c$ is a scaling constant that adjusts the weight of the misclassification function $f$, which is designed to increase the likelihood of misclassification:
     \begin{equation}
        f(\mathbf{F}_k') = \max(\max_{i \neq t} Z(\mathbf{F}_k')_i - Z(\mathbf{F}_k')_t, -\kappa),
    \end{equation}
    where $Z(\mathbf{F}_k')$ outputs the logits from the model for the perturbed input, with $t$ indicating the target class, and $\kappa$ serving as a confidence parameter to ensure robustness in the adversarial example.

    \item \textbf{Basic Iterative Method (BIM) Attack} \citep{bim}: The BIM attack incrementally adjusts an initial input by applying small but cumulative perturbations, based on the sign of the gradient of the loss function with respect to the input, aiming to maximize the prediction error in a model while ensuring the perturbations remain within specified bounds:
    \begin{equation}
    \mathbf{F}_k^{t+1} = \text{Clip}_{\Delta}\{\mathbf{F}_k^t + \alpha \cdot \text{sign}(\nabla{\mathbf{F}_k^t} \mathcal{L}(\mathbf{F}_k^t, \mathbf{y}))\}, 
    \end{equation}
    where $t$ represents the iteration index, $\alpha$ is the size of the step, $\Delta$ defines the maximum allowable perturbation, and $\text{Clip}_{\Delta}$ is a function that restricts the values within a $\Delta$ boundary around the original features $\mathbf{F}_k$. The initial setting is $\mathbf{F}_k^0 = \mathbf{F}_k$. This procedure is iterated either a preset number of times or until a specific stopping condition is reached.
\end{itemize}